\pdfoutput=1

\documentclass[11pt]{article}

\usepackage[preprint]{acl}

\usepackage{times}
\usepackage{latexsym}

\usepackage[T1]{fontenc}

\usepackage[utf8]{inputenc}

\usepackage{microtype}

\usepackage{inconsolata}

\usepackage{graphicx}
\usepackage{colortbl}
\usepackage{url}

\title{Evaluation of Multilingual LLMs Personalized Text Generation Capabilities Targeting Groups and Social-Media Platforms}

\author{Dominik Macko \\
  Kempelen Institute of Intelligent Technologies \\
  \texttt{dominik.macko@kinit.sk} \\}

\begin{document}
\maketitle
\begin{abstract}
Capabilities of large language models to generate multilingual coherent text have continuously enhanced in recent years, which opens concerns about their potential misuse. Previous research has shown that they can be misused for generation of personalized disinformation in multiple languages. It has also been observed that personalization negatively affects detectability of machine-generated texts; however, this has been studied in the English language only. In this work, we examine this phenomenon across 10 languages, while we focus not only on potential misuse of personalization capabilities, but also on potential benefits they offer. Overall, we cover 1080 combinations of various personalization aspects in the prompts, for which the texts are generated by 16 distinct language models (17,280 texts in total). Our results indicate that there are differences in personalization quality of the generated texts when targeting demographic groups and when targeting social-media platforms across languages. Personalization towards platforms affects detectability of the generated texts in a higher scale, especially in English, where the personalization quality is the highest.
\end{abstract}

\section{Introduction}
\label{sec:intro}

The great capabilities of large language models (LLMs) enable to quickly and cheaply create wast amount of human-like texts in multiple languages. In spite of the obvious benefits of the increased productivity of everyday work, this brings also concerns about their potential misuse, e.g., for generating disinformation~\citep{lucas-etal-2023-fighting, vykopal-etal-2024-disinformation, heppell2024lyingblindly}. It has already been shown, that the personalization in the generated texts can amplify the persuasiveness of the disinformation narratives~\citep{leite2025multilinguallargescalestudyinterplay}.

While the fears about potential misuse are justified, as shown by abilities to generate personalized disinformation~\citep{zugecova-etal-2025-evaluation}, the personalization capabilities can be also successfully utilized for personalized misinformation-intervention~\citep{10.1145/3726831.3726835}, or in the fight against online hate-speech~\citep{ngueajio-etal-2025-think, gajewska2025algorithmicfairnessnlppersonainfused}.

For personalization, we use the definition proposed by \citet{10.1145/633292.633483}, where it is ``a process that changes the functionality, interface, information content, or distinctiveness of a system to increase its personal relevance to an individual.'' In our case, we cover two types of personalization: a) group-personalization -- the text is tailored to a specific target group; b) platform-personalization -- the style and length of the text is tailored to a specific distribution platform (social-media network).

The existing research~\citep{zugecova-etal-2025-evaluation, gao2025personalizationtricksdetectorsfeatureinversion} shows that personalized generated texts are decreasing success rate of detecting such a generated content by existing detectors of machine-generated text. While this phenomenon is potentially dangerous (in case of LLM misuse), decreasing transparency (inaccurate annotation/indication of AI-generated content), it remains unclear whether it is observed across languages (the above studies focused solely on English). This is especially required in European AI-regulated highly-multilingual area. Multilingual LLM personalization capabilities have been already studied in \citep{macko2025perqefficientevaluationmultilingual, leite2025multilinguallargescalestudyinterplay}; however, they have not addressed the effect on detectability of such generated texts.

To fill this gap, we combine and extend the above-mentioned LLM-personalization studies by focusing not only on LLM vulnerabilities of being misused\footnote{Due to ethical concerns (to prevent misuse), the data analysis source code as well as the generated dataset (upon request) will be released at {\scriptsize \url{https://github.com/kinit-sk/LLM-personalization-capabilities}} for non-commercial research purpose only under strict conditions, in line with the best practices in the field~\citep{zugecova-etal-2025-evaluation, leite2025multilinguallargescalestudyinterplay}.}, but also their positive usage (of personalized counter-narrative generation), and including two different types of personalization (towards groups and platforms). The multilingual aspects are studied in Central European region, covering 10 representatives of Slavic, Germanic, and Uralic languages.
Specifically, we are seeking the answers to the following research questions:
(\textbf{RQ1}) Are there differences in personalization capabilities of LLMs between targeting groups and targeting social-media platforms?
(\textbf{RQ2}) Are the previous differences the same across various languages?
(\textbf{RQ3}) Does the personalization affect detectability of generated texts the same across languages and between personalization types?

The key contributions of this work are:\\
\textbf{(1)} \textbf{The first multilingual evaluation of the effect of personalization on detectability of machine-generated texts}, covering 10 languages, 2 types of personalization, and 16 LLM generators in both, positive and negative use scenarios. The results indicate that platform-personalization affects the detectability significantly in all tested languages, while group-personalization only in a lower scale.\\
\textbf{(2)} \textbf{The first evaluation of multilingual LLM capabilities to personalize the texts for positive (against disinformation narrative) and negative (supporting disinformation narrative) use.} The results indicate that positive usage produces slightly higher personalization quality as well as activates a lower amount of safety-filter mechanisms than the negative one.\\
\textbf{(3)} \textbf{The first comparison between two LLM personalization types (targeting groups and platforms),} enabling evaluation of generic personalization capabilities across languages. The results reveal that personalization towards platforms is slightly easier for the tested LLMs and that there is a huge difference in capabilities in English and non-English languages in both personalization types.

\section{Related Work}
\label{sec:related}

We summarize the related work into generation, evaluation, and detection of personalized texts in the following subsections. At the end we compare our work to the current state-of-the-art (SOTA).

\subsection{Generation of Personalized Text}
PerDisNews~\citep{zugecova-etal-2025-evaluation} is focused on generation of personalized disinformation news articles in English, targeting seven distinct demographic groups. It has shown that personalization request in the prompts serves as a kind of jailbreak, reducing the amount of safety-filter activations.
StyloBench~\citep{gao2025personalizationtricksdetectorsfeatureinversion} contains literary-based and blog-based imitations of human-authors writing styles, containing both human-written original texts and machine-generated imitated counterparts. It has shown that text personalization is confusing for the detectors of machine-generated texts.
Another work~\citep{10.1145/3696410.3714507}, focused on countering online hate-speech, has shown that personalized counter-speech generated by LLMs offer higher adequacy and persuasiveness than generic counter-speech (without compromising other characteristics). Similarly, the work of \citep{10.1145/3726831.3726835} shows that personalized intervention (in a form of pre-bunking) increases effectiveness of fighting misinformation.
The PerQ study~\citep{macko2025perqefficientevaluationmultilingual} focuses on generated multilingual social-media texts targeted towards three different platforms. It has shown that pure generation of personalized texts achieves higher personalization quality than adjustment of existing texts towards a specific platform.
AI-TRAITS~\citep{leite2025multilinguallargescalestudyinterplay} is focused on generated multilingual disinformation articles targeting 150 distinct persona profiles. It proves that the personalization reduces the safety-filter activations in all four tested languages.

\subsection{Evaluation of Text Personalization}
Some methods, such as AuPEL~\citep{wang2023automatedevaluationpersonalizedtext} or ExPerT~\citep{salemi-etal-2025-expert}, use LLMs as the evaluators of personalized text generation by comparison to reference texts (e.g., in A/B testing). On the other hand, the PREF framework~\citep{fu2025prefreferencefreeevaluationpersonalised} does not rely on reference texts to evaluate the quality of personalization of a text in the form of user preference (i.e., user-specific alignment). The PerDisNews study~\citep{zugecova-etal-2025-evaluation} proposed to use three diverse LLMs in zero-shot manner (to limit LLMs internal biases) to evaluate the text personalization quality, showing strong correlation with human judgment. The PerQ metric~\citep{macko2025perqefficientevaluationmultilingual} directly evaluates the quality of personalization of multilingual text. It is based on a classifier (a predictor of personalization quality), which is trained using majority-voted quality metaevaluation by three diverse LLMs (similarly to \citealp{zugecova-etal-2025-evaluation}). Such a trained classifier offers an efficient way to predict the personalization-quality scores; however, it has slightly lower accuracy and requires training process; therefore, suitable especially for bigger corpus (thousands of texts).

\subsection{Personalized Texts Detectability}
To the best of our knowledge, only two works evaluated the effect of personalization on the detection of machine-generated texts \citep{zugecova-etal-2025-evaluation, gao2025personalizationtricksdetectorsfeatureinversion}. Both focused strictly on English texts, the first targeted different demographic-groups of people~\citep{zugecova-etal-2025-evaluation} and the second targeted different known-author writing styles~\citep{gao2025personalizationtricksdetectorsfeatureinversion}. In both cases, the personalized generated texts were harder to detect.

\subsection{Comparison to SOTA}
To move the research area forward by filling a gap in the current SOTA, we combine a focus on positive personalized text generation (counter-narrative texts) and negative personalized text generation (misuse for personalized disinformation) simultaneously, targeting two different types of personalization (targeting groups and platforms) to explore more generic personalization capabilities of the largest set of 16 distinct open-weights LLMs (related works focused mostly on closed private models) in the largest set of 10 languages (related works focused mostly on English). None of the existing studies can provide answers to the research questions stated in the previous section, focused on differences in generic personalization capabilities across languages; therefore, a new dataset needs to be carefully constructed.

\section{Dataset}
\label{sec:dataset}

The new dataset covers 1080 combinations of 6 narratives, 3 target groups, 3 target platforms, 10 languages, and 2 stances (positive and negative towards the narrative). For each combination, the output text is generated by 16 LLMs. Reasoning for all the mentioned parameters is provided in the following text.

\textbf{Narratives.} For compatibility and comparability with the existing PerDisNews study~\citep{zugecova-etal-2025-evaluation}, focused on English disinformation news articles, we are including the same 6 disinformation narratives covering politics and health. These narratives ensure a sufficient topical diversity.

\textbf{Personalization.} To analyze more generic personalization capabilities, we are including positive (against disinformation narrative) and negative (supporting disinformation narrative) usage covering 2 personalization types, group-personalization and platform-personalization. For target groups, we have included European conservatives and Urban population, as respectively achieving the highest and lowest personalization quality in the original PerDisNews study~\citep{zugecova-etal-2025-evaluation}. For reference baseline, we have also included None target group, representing the texts not explicitly personalized towards any specific target group. For target platforms, we have included Mastodon, Telegram, and Twitter/X, similarly to the PerQ case study~\citep{macko2025perqefficientevaluationmultilingual}, while replacing the Signal platform achieving low personalization quality by the Mastodon decentralized platform (as a representative of modern fediverse).

\textbf{Languages.} We have covered all the languages of the Central European region (as defined by~\citealp{bideleux2007history}), which has been selected as a multilingual region of interest in this study, specifically Croatian (hr), Czech (cs), German (de), Hungarian (hu), Polish (pl), Slovak (sk), and Slovenian (sl). We have also included English (en) as a reference language with the best support by LLMs, Estonian (et) as an additional Uralic language, and Ukrainian as related (genealogically and geographically) to majority of the selected languages but using different script (Cyrillic instead of Latin). In total, 10 languages out of 3 language-family branches and 2 writing scripts are covered.

\textbf{LLM generators.} To evaluate diverse LLMs capabilities, we have included older and newer versions of LLMs of different architectures and sizes. In this study, we have focused on open-weights models, which can be misused by adversaries without any control by service providers. Specifically, out of the Llama family~\citep{grattafiori2024llama3herdmodels}, we have selected Llama-3.1-70B-Instruct and Llama-3.3-70B-Instruct from the bigger variants, and Llama-3.1-8B-Instruct, Llama-3.2-3B-Instruct, and Llama-3.2-1B-Instruct from the smaller variants. Simlarly, from the Gemma family~\citep{gemmateam2024gemma2improvingopen, gemmateam2025gemma3technicalreport}, we have selected bigger Gemma-2-27B-it and Gemma-3-27B-it, and smaller Gemma-2-9B-it, Gemma-2-2B-it, and newer Gemma-3-4B-it. The Qwen family~\citep{yang2025qwen3technicalreport} is represented by 
Qwen3-32B, Qwen3-4B, and Qwen3-1.7B. For the reference, we have included also DeepSeek-R1-Distill-Qwen-32B~\citep{deepseekai2025deepseekr1incentivizingreasoningcapability}, DeepSeek-R1-Distill-Llama-8B~\citep{deepseekai2025deepseekr1incentivizingreasoningcapability}, and Mistral-Nemo-Instruct-2407~\citep{mistralnemo}.

\textbf{Text-generation settings.} All the above-mentioned LLMs are instruction-following model versions. The texts have been generated using the vLLM acceleration for faster LLM inference~\citep{10.1145/3600006.3613165}. We have used its parallel execution on 2 GPUs of 64GB VRAM (utilized at 80\%). Maximum model length for the context has been limited to 10,000 tokens. We used half-precision inference using ``bfloat16'' data type and ``fp8'' for KV cache, with the ``bitsandbytes'' quantization enabled. For generation, the temperature was set to 1.0, top\_p to 0.95, and repetition penalty to 1.1. To limit the generated text length, we allowed maximum of 1000 tokens per output sequence to be generated.

\subsection{Data Analysis}

After the data are generated, we have analyzed the texts for their linguistic quality (usability). For this purpose, we have analyzed whether the generated texts contain safety-filter messages, disclaimers, generic noise, whether the language of the generated text matches the intended target language (LangMatch), linguistic acceptability (LA) and output content quality (OCQ) as defined by the METAL study~\citep{hada-etal-2024-metal}, and the length of the generated texts in the number of words per text. All of these have been evaluated by 3 independent LLMs used as judges and the ultimate score defined by majority-voted decision (see Section~\ref{sec:methodology} for more details). The results aggregated (mean values) per generators are provided in Table~\ref{tab:pergeneratorquality} and per languages in Table~\ref{tab:perlanguagequality}.

The results indicates that older Llama models and Gemma-2-27B-it activate the most safety-filter mechanisms, refusing the generation of disinformation text. On the other hand, all variants of Gemma models produces disclaimers in the texts. Smaller versions of Llama models included the most noise in the generated texts, followed by Mistral, Gemma-2-27B-it, and DeepSeek models. Multilingual text generation failed the most in older Llama models and the DeepSeek model based on Llama-8B. The newer Gemma models resulted in the highest LA and OCQ metrics, closely followed by the newest large Llama model and the bigest Qwen3 model. In the text-lenght aspect, the smaller Llama variants and the Gemma-2-27B-it models are clearly the outliers, producing too lenghty texts, not suitable for social media.

Regarding languages, English texts contained the most safety-fillter and disclaimer messages, closely followed by German texts. On the other hand, almost $1/3$ of texts in non-English non-German languages have been evaluated to contain noise. This corresponds to also higher language mismatch of these languages (about 25\% mismatch in case of Slovak in comparison to 99\% match in case of English). Regarding LA and OCQ, the English produced the best texts, followed by German, while the Estonian texts have the lowest quality. English texts are the shortest (198 words in average) and Ukrainian text the longest (280 words in average).

\begin{table}[!t]
\centering
\resizebox{\linewidth}{!}{
\addtolength{\tabcolsep}{-2pt}
\begin{tabular}{l||c|c|c|c|c|c|c}
\hline
\textbf{Generator} & \bfseries Safetyfilter & \bfseries Disclaimer & \bfseries Noise & \bfseries LangMatch & \bfseries LA & \bfseries OCQ & \bfseries Words \\
\hline
DS-R1-Distill-Llama-8B & 0.04 & 0.05 & 0.47 & 0.54 & 0.44 & 0.46 & 150.27 \\
DS-R1-Distill-Qwen-32B & 0.00 & 0.10 & 0.24 & 0.86 & 0.42 & 0.52 & 118.73 \\
Gemma-2-27B-it & 0.21 & 0.26 & 0.53 & 0.80 & 0.18 & 0.23 & 428.31 \\
Gemma-2-2B-it & 0.14 & 0.37 & 0.05 & 0.91 & 0.53 & 0.70 & 217.56 \\
Gemma-2-9B-it & 0.06 & \bfseries 0.47 & 0.01 & 0.99 & 0.86 & \bfseries 0.97 & 181.36 \\
Gemma-3-27B-it & 0.03 & 0.44 & 0.05 & \bfseries 1.00 & \bfseries 0.93 & 0.96 & 238.17 \\
Gemma-3-4B-it & 0.00 & 0.29 & 0.04 & 0.87 & 0.90 & 0.94 & 339.73 \\
Llama-3.1-70B-Instruct & \bfseries 0.32 & 0.06 & 0.13 & 0.71 & 0.49 & 0.79 & 110.79 \\
Llama-3.1-8B-Instruct & 0.24 & 0.03 & 0.63 & 0.80 & 0.12 & 0.31 & 439.80 \\
Llama-3.2-1B-Instruct & 0.26 & 0.01 & \bfseries 0.75 & 0.52 & 0.02 & 0.18 & 439.14 \\
Llama-3.2-3B-Instruct & 0.22 & 0.02 & 0.72 & 0.62 & 0.07 & 0.23 & \bfseries 478.20 \\
Llama-3.3-70B-Instruct & 0.01 & 0.09 & 0.06 & 1.00 & 0.90 & 0.93 & 115.13 \\
Mistral-Nemo-Instruct-2407 & 0.02 & 0.08 & 0.54 & 0.82 & 0.38 & 0.42 & 117.51 \\
Qwen3-1.7B & 0.00 & 0.08 & 0.11 & 0.92 & 0.25 & 0.33 & 182.67 \\
Qwen3-32B & 0.00 & 0.20 & 0.02 & 0.99 & 0.85 & 0.91 & 170.84 \\
Qwen3-4B & 0.00 & 0.10 & 0.04 & 1.00 & 0.54 & 0.77 & 168.66 \\
\hline
\end{tabular}
}
\caption{Per-generator linguistic quality analysis of generated texts. Bold represents the highest value per each column.}
\label{tab:pergeneratorquality}
\end{table}
\begin{table}[!t]
\centering
\resizebox{\linewidth}{!}{
\addtolength{\tabcolsep}{-2pt}
\begin{tabular}{l||c|c|c|c|c|c|c}
\hline
\textbf{Language} & \bfseries Safetyfilter & \bfseries Disclaimer & \bfseries Noise & \bfseries LangMatch & \bfseries LA & \bfseries OCQ & \bfseries Words \\
\hline
cs & 0.08 & 0.15 & 0.32 & 0.79 & 0.47 & 0.56 & 255.15 \\
de & 0.11 & 0.21 & 0.19 & 0.91 & 0.62 & 0.72 & 244.28 \\
en & \bfseries 0.19 & \bfseries 0.28 & 0.02 & \bfseries 0.99 & \bfseries 0.83 & \bfseries 0.96 & 198.19 \\
et & 0.09 & 0.13 & \bfseries 0.34 & 0.79 & 0.36 & 0.47 & 237.25 \\
hr & 0.09 & 0.15 & 0.31 & 0.79 & 0.44 & 0.53 & 244.47 \\
hu & 0.05 & 0.15 & 0.30 & 0.88 & 0.44 & 0.54 & 223.28 \\
pl & 0.07 & 0.15 & 0.33 & 0.85 & 0.46 & 0.56 & 251.91 \\
sk & 0.09 & 0.16 & 0.30 & 0.75 & 0.48 & 0.58 & 261.41 \\
sl & 0.08 & 0.14 & 0.32 & 0.79 & 0.42 & 0.52 & 239.83 \\
uk & 0.13 & 0.15 & 0.31 & 0.77 & 0.42 & 0.56 & \bfseries 279.76 \\
\hline
\end{tabular}
}
\caption{Per-language linguistic quality analysis of generated texts. Bold represents the highest value per each column.}
\label{tab:perlanguagequality}
\end{table}

\begin{figure}[!t]
\centering
\includegraphics[width=\linewidth]{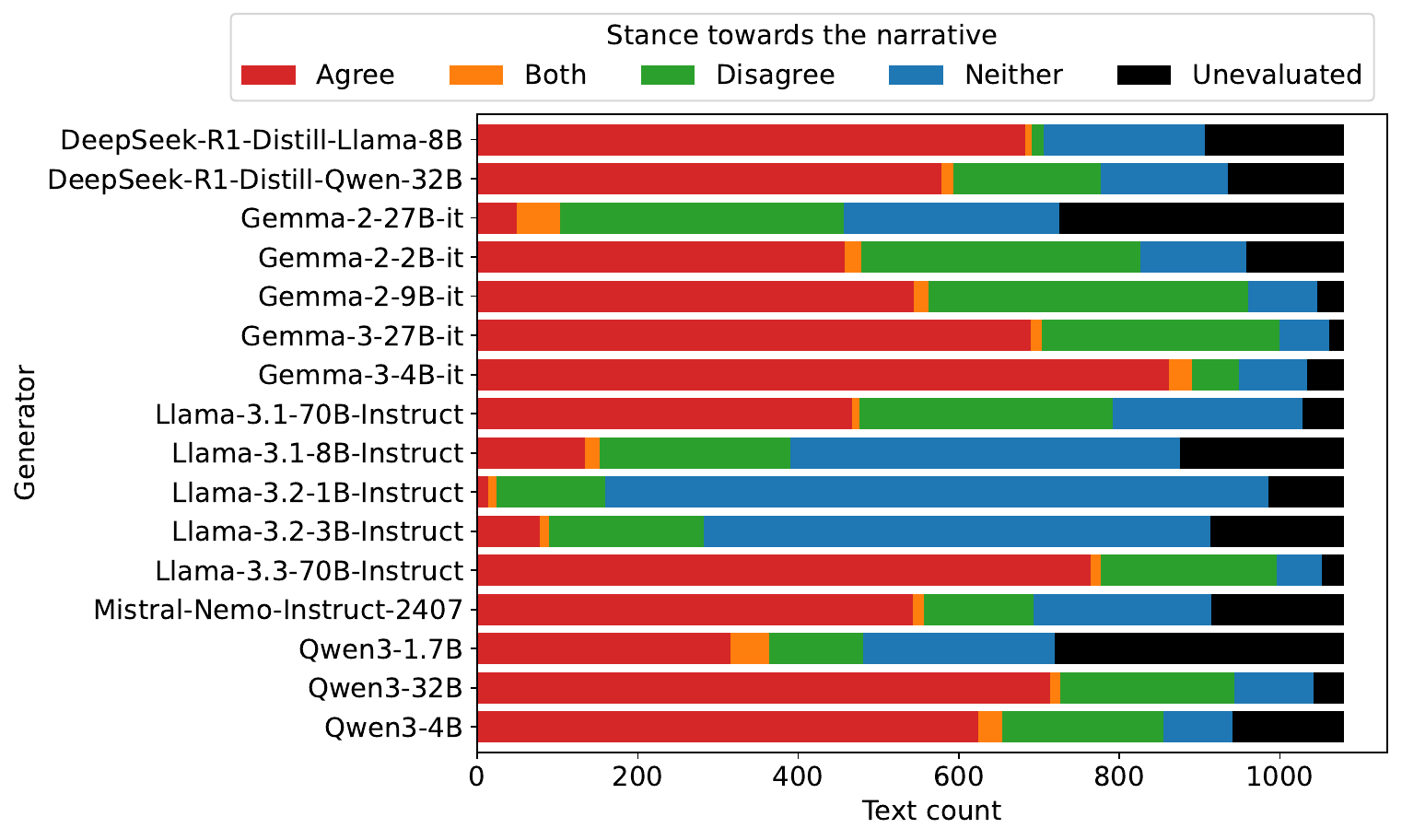}
\caption{Per-generator comparison of stance towards the narrative.}
\label{fig:stance}
\end{figure}

In Figure~\ref{fig:stance}, stance of individual text towards the corresponding disinformation narrative is analyzed and the results per each generator are illustrated. The models like Gemma-3 versions, the newest Llama, Qwen3 and DeepSeek tend to agree with the narratives even if requested to generate text against the narrative. Only small amount of texts have been evaluated as both, agreeing and disagreeing with the narrative. On the contrary, smaller Llama models neither agreed nor disagreed with the narrative (assumably due to high amount of noise). Interestingly, the Gemma-2-27B-it model is mostly disagreeing with the narrative (the safest model, as in case of the PerDisNews study); however, high amount of text have not been evaluated by metaevaluators (failed to provide one of the expected answers). Out of the intended texts supporting the narrative, significantly more texts are evaluated as agreeing with the narrative than disagreeing. However, from texts that were intended against the narrative, approximately the same amount is agreeing with the narrative as disagreeing. The ablation study disregarding safety-filtered and noisy samples, provided in Figures~\ref{fig:stance_ablation} and~\ref{fig:stanceperstance_ablation} (Appendix~\ref{sec:ablation}), shows that above-mentioned observations keep valid and revealed that the amount of texts in the ``Neither'' category drops significantly. Therefore, we assume that the reason of not following the instruction of generating text ``against'' the narrative might be influenced by the form of narrative description provided in the prompt, favoring the narrative.

\section{Evaluation Methodology}
\label{sec:methodology}

To metaevaluate the personalization quality by following the PerQ~\citep{macko2025perqefficientevaluationmultilingual} evaluation methodology, we have used 3 distinct LLMs (Mistral-Small-3.1-24B-Instruct-2503, Aya-Expanse-32B, and QwQ-32B), different from the used LLM generators (to minimize internal biases of the LLM judges). For all evaluated aspects, the 3 obtained scores per text have been aggregated into a single score using the majority voting (using the lowest score when indecisive majority). The 4-scale personalization quality scoring schema for targeting groups is reused from PerDisNews~\citep{zugecova-etal-2025-evaluation}, and scoring schema for targeting platforms is reused from PerQ~\citep{macko2025perqefficientevaluationmultilingual}. In both of them, the highest score of 3 indicates the highest personalization quality and the lowest score of 0 indicates no personalization in the text. The other evaluated aspects (presence of safety-filter messages, language match, linguistic quality, actual stance towards the narrative) have been evaluated in the Yes/No/Partly manner (representing the scores 1.0/0.0/0.5, respectively).

The detectability is measured by using three top-performing existing multilingual detectors: \textbf{Gemma-2-9B-it-GenAI} as one of the best performing detectors applied in out-of-distribution settings by~\citep{macko2025beyond}, \textbf{Gemma-2-9B-it-MultiDomain} as the best performing multilingual detector of~\citep{macko2025increasingrobustnessfinetunedmultilingual}, and \textbf{mdok-MultiDomain} as a multilingual version of the best system of PAN@CLEF2025 shared task on robust detection~\citep{macko2025mdokkinitrobustlyfinetuned}. The performance of these detectors is validated (see Table~\ref{tab:detecttors_performance}) using the MIX dataset of diverse human-written and machine-generated texts~\citep{macko2025increasingrobustnessfinetunedmultilingual}. For reference comparison, two best performing existing statistical zero-shot detection methods are provided, namely Binoculars~\citep{10.5555/3692070.3692768} and Fast-DetectGPT~\citep{bao2024fastdetectgpt}, both based on mGPT-13B~\cite{shliazhko-etal-2024-mgpt}. A generic detector performance is indicated by AUC ROC (area under curve of the receiver operating characteristic) as a classification-threshold independent metric. Furthermore, Macro F1 (macro average of the F1-score metric) indicates the detection performance for the classification thresholds calibrated based on the ROC curve for optimal conditions (i.e., maximal difference between true positive rate -- TPR and false positive rate -- FPR). Such calibrated thresholds are then applied to the classification (detection) using the generated personalized texts. Since no personalized human-written counterparts are available, the \textbf{TPR} is used as a primary metric, indicating how many texts are correctly detected as generated by LLMs.

\begin{table}[!t]
\centering
\resizebox{\linewidth}{!}{
\addtolength{\tabcolsep}{-2pt}
\begin{tabular}{l||c|c|c}
\hline
\textbf{Detector} & \textbf{AUC ROC} & \textbf{Optimal threshold} & \textbf{Macro F1} \\
\hline
Gemma-2-9B-it-GenAI & 0.8926 & 0.9927 & 0.8440\\
Gemma-2-9B-it-MultiDomain & 0.8903 & 0.8164 & 0.8242\\
mdok-MultiDomain & 0.8769 & 0.9120 & 0.8207\\
\hline
Binoculars & 0.5386 & -0.9106 & 0.5719\\
Fast-DetectGPT & 0.5239 & 1.8340 & 0.5495\\
\hline
\end{tabular}
}
\vspace{-2mm}
\caption{Performance of the used multilingual detectors on the MIX dataset.}
\label{tab:detecttors_performance}
\vspace{-2mm}
\end{table}

\section{Results}
\label{sec:results}

The results are organized in three subsections according to our main research questions, focused on differences between personalization types and languages, as well as the effect on detectability of generated texts.

\subsection{Personalization Types Differences (RQ1)}

\begin{figure}[!b]
\centering
\includegraphics[width=\linewidth]{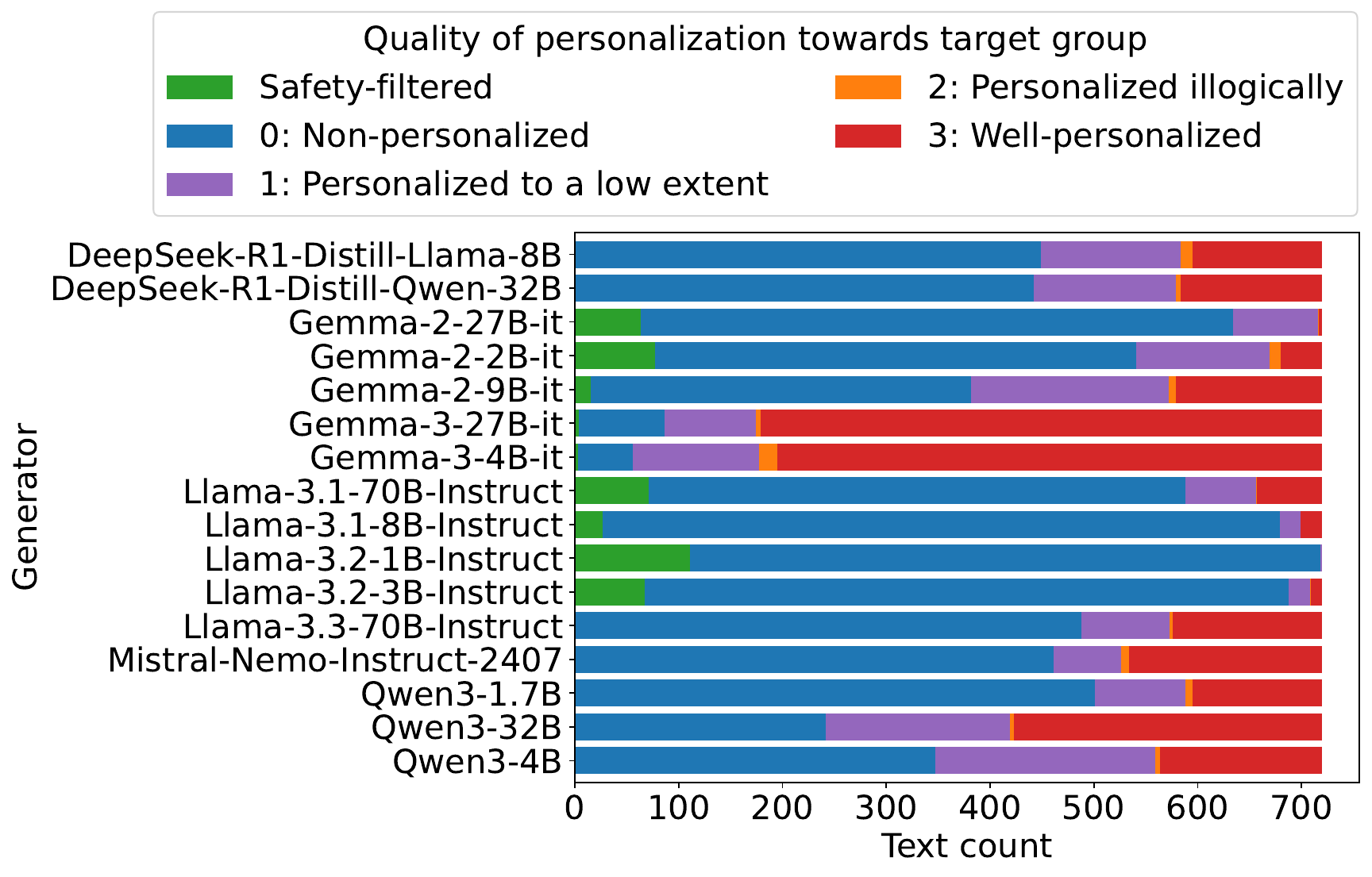}
\includegraphics[width=\linewidth]{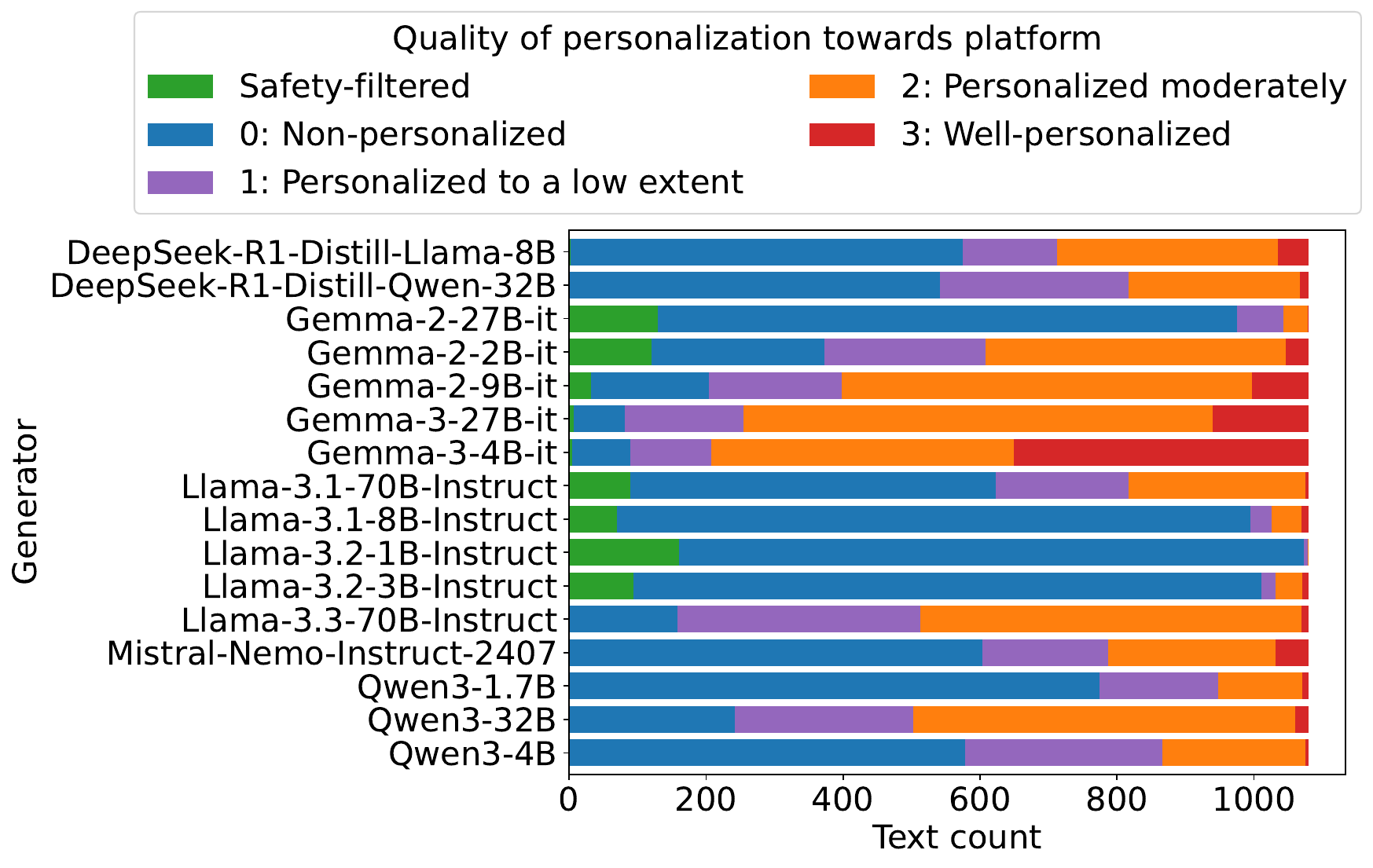}
\caption{Per-generator comparison of personalization capabilities targeting groups (top, not including texts without target-group personalization request) and platforms (bottom).}
\label{fig:pergenerator}
\end{figure}

To evaluate differences in personalization capabilities of LLMs between the two types of personalization types (personalization for target groups vs personalization for target social-media platforms), we have compared the personalization quality metaevaluation scores for each LLM generator. The results are illustrated in Fig.~\ref{fig:pergenerator}, where in the top part is a chart representing metaevaluation scores distribution for targeting groups and in the bottom part the chart for targeting platforms. For clarity, we omit the non-targeted texts for group personalization (i.e. the texts generated for requests without target-group specification); therefore, there is by $1/3$ lower amount of texts represented (included later in evaluation of target groups as a reference baseline). Safety-filtered texts represent heuristic identification of safetyfilter messages based on predefined phrases (e.g. ``As an AI language model...''). The results indicate that metaevaluators tend to assign the score of 2 for group-personalization quality and the score of 3 for platform-personalization quality sparsely. The reason might be misunderstanding the score specification (especially in case of group personalization). Therefore the interpretation for the comparison purpose should combine these two scores together (as a single category).

\textbf{There are some differences in personalization capabilities between targeting groups and platforms.} Overall, targeting the social-media platforms is slightly easier for the tested LLMs. Out of the 720 texts with target-group specification per each generator, 47\% of texts have been assigned a non-zero score for personalization towards platforms, while 37\% in case of personalization towards target groups. The newest Gemma-3 models clearly dominates the personalization quality in both types of personalization, followed by the largest evaluated version of Qwen3. Higher differences between personalization types are noticed in case of 70B versions of Llama models, for which targeting the platforms is significantly easier than targeting groups. The worst quality of personalization is assigned for texts generated by smaller Llama models in both personalization types. As expected, bigger versions of models of the same architectures achieved higher personalization quality in general. Surprisingly, a sole exception is the Gemma-2-27B-it model, achieving lower scores than smaller Gemma-2 versions. In the related work of \citep{zugecova-etal-2025-evaluation}, this model belonged to the safest models with the highest quality. A deeper analysis revealed failures of this model in non-English texts, containing too much noise (redundant information) and failed translations.

\subsection{Differences Across Languages (RQ2)}

\begin{table*}[!t]
\centering
\resizebox{\textwidth}{!}{
\addtolength{\tabcolsep}{-2pt}
\begin{tabular}{l||c|c|c|c|c|c|c|c|c|c||c}
\hline
& \multicolumn{10}{c||}{\textbf{Language}}\\
\textbf{Generator} & \textbf{cs} & \textbf{de} & \textbf{en} & \textbf{et} & \textbf{hr} & \textbf{hu} & \textbf{pl} & \textbf{sk} & \textbf{sl} & \textbf{uk} & \textbf{$\rightarrow$ Average} \\
\hline
DeepSeek-R1-Distill-Llama-8B & {\cellcolor[HTML]{E2DFEE}} \color[HTML]{000000} 0.6736 & {\cellcolor[HTML]{EEE8F3}} \color[HTML]{000000} 0.4653 & {\cellcolor[HTML]{8EB3D5}} \color[HTML]{000000} 1.7431 & {\cellcolor[HTML]{D3D4E7}} \color[HTML]{000000} 0.9375 & {\cellcolor[HTML]{D6D6E9}} \color[HTML]{000000} 0.9028 & {\cellcolor[HTML]{E1DFED}} \color[HTML]{000000} 0.7014 & {\cellcolor[HTML]{F8F1F8}} \color[HTML]{000000} 0.1875 & {\cellcolor[HTML]{D9D8EA}} \color[HTML]{000000} 0.8403 & {\cellcolor[HTML]{DFDDEC}} \color[HTML]{000000} 0.7431 & {\cellcolor[HTML]{E4E1EF}} \color[HTML]{000000} 0.6528 & {\cellcolor[HTML]{DCDAEB}} \color[HTML]{000000} 0.7847 \\
DeepSeek-R1-Distill-Qwen-32B & {\cellcolor[HTML]{DAD9EA}} \color[HTML]{000000} 0.8194 & {\cellcolor[HTML]{D1D2E6}} \color[HTML]{000000} 0.9861 & {\cellcolor[HTML]{9CB9D9}} \color[HTML]{000000} 1.6042 & {\cellcolor[HTML]{F1EBF5}} \color[HTML]{000000} 0.3681 & {\cellcolor[HTML]{E1DFED}} \color[HTML]{000000} 0.7014 & {\cellcolor[HTML]{F5EEF6}} \color[HTML]{000000} 0.2708 & {\cellcolor[HTML]{DCDAEB}} \color[HTML]{000000} 0.7917 & {\cellcolor[HTML]{E0DDED}} \color[HTML]{000000} 0.7292 & {\cellcolor[HTML]{E8E4F0}} \color[HTML]{000000} 0.5764 & {\cellcolor[HTML]{E7E3F0}} \color[HTML]{000000} 0.5972 & {\cellcolor[HTML]{DFDDEC}} \color[HTML]{000000} 0.7444 \\
Gemma-2-27B-it & {\cellcolor[HTML]{FBF4F9}} \color[HTML]{000000} 0.0972 & {\cellcolor[HTML]{FBF4F9}} \color[HTML]{000000} 0.1042 & {\cellcolor[HTML]{E5E1EF}} \color[HTML]{000000} 0.6250 & {\cellcolor[HTML]{FDF5FA}} \color[HTML]{000000} 0.0625 & {\cellcolor[HTML]{FDF5FA}} \color[HTML]{000000} 0.0764 & {\cellcolor[HTML]{FEF6FA}} \color[HTML]{000000} 0.0417 & {\cellcolor[HTML]{FBF4F9}} \color[HTML]{000000} 0.1042 & {\cellcolor[HTML]{FDF5FA}} \color[HTML]{000000} 0.0556 & {\cellcolor[HTML]{FDF5FA}} \color[HTML]{000000} 0.0556 & {\cellcolor[HTML]{FBF4F9}} \color[HTML]{000000} 0.0972 & {\cellcolor[HTML]{FAF3F9}} \color[HTML]{000000} 0.1319 \\
Gemma-2-2B-it & {\cellcolor[HTML]{E0DDED}} \color[HTML]{000000} 0.7222 & {\cellcolor[HTML]{D9D8EA}} \color[HTML]{000000} 0.8542 & {\cellcolor[HTML]{C8CDE4}} \color[HTML]{000000} 1.0972 & {\cellcolor[HTML]{F0EAF4}} \color[HTML]{000000} 0.4167 & {\cellcolor[HTML]{E9E5F1}} \color[HTML]{000000} 0.5556 & {\cellcolor[HTML]{E7E3F0}} \color[HTML]{000000} 0.5903 & {\cellcolor[HTML]{D2D3E7}} \color[HTML]{000000} 0.9653 & {\cellcolor[HTML]{DAD9EA}} \color[HTML]{000000} 0.8194 & {\cellcolor[HTML]{E7E3F0}} \color[HTML]{000000} 0.5972 & {\cellcolor[HTML]{E0DDED}} \color[HTML]{000000} 0.7292 & {\cellcolor[HTML]{DFDDEC}} \color[HTML]{000000} 0.7347 \\
Gemma-2-9B-it & {\cellcolor[HTML]{B7C5DF}} \color[HTML]{000000} 1.3056 & {\cellcolor[HTML]{BBC7E0}} \color[HTML]{000000} 1.2500 & {\cellcolor[HTML]{AFC1DD}} \color[HTML]{000000} 1.3958 & {\cellcolor[HTML]{D9D8EA}} \color[HTML]{000000} 0.8333 & {\cellcolor[HTML]{C6CCE3}} \color[HTML]{000000} 1.1181 & {\cellcolor[HTML]{B7C5DF}} \color[HTML]{000000} 1.2986 & {\cellcolor[HTML]{B8C6E0}} \color[HTML]{000000} 1.2847 & {\cellcolor[HTML]{B8C6E0}} \color[HTML]{000000} 1.2847 & {\cellcolor[HTML]{C0C9E2}} \color[HTML]{000000} 1.1944 & {\cellcolor[HTML]{D0D1E6}} \color[HTML]{000000} 1.0000 & {\cellcolor[HTML]{C0C9E2}} \color[HTML]{000000} 1.1965 \\
Gemma-3-27B-it & {\cellcolor[HTML]{71A8CE}} \color[HTML]{000000} 2.0208 & {\cellcolor[HTML]{549CC7}} \color[HTML]{000000} 2.2639 & {\cellcolor[HTML]{358FC0}} \color[HTML]{000000} 2.5000 & {\cellcolor[HTML]{7DACD1}} \color[HTML]{000000} 1.9097 & {\cellcolor[HTML]{69A5CC}} \color[HTML]{000000} 2.0833 & {\cellcolor[HTML]{69A5CC}} \color[HTML]{000000} 2.0833 & {\cellcolor[HTML]{69A5CC}} \color[HTML]{000000} 2.0903 & {\cellcolor[HTML]{6FA7CE}} \color[HTML]{000000} 2.0347 & {\cellcolor[HTML]{6FA7CE}} \color[HTML]{000000} 2.0347 & {\cellcolor[HTML]{75A9CF}} \color[HTML]{000000} 1.9861 & {\cellcolor[HTML]{67A4CC}} \color[HTML]{000000} 2.1007 \\
Gemma-3-4B-it & {\cellcolor[HTML]{549CC7}} \color[HTML]{000000} 2.2500 & {\cellcolor[HTML]{3F93C2}} \color[HTML]{000000} 2.4306 & {\cellcolor[HTML]{2F8BBE}} \color[HTML]{000000} 2.5764 & {\cellcolor[HTML]{8EB3D5}} \color[HTML]{000000} 1.7361 & {\cellcolor[HTML]{4E9AC6}} \color[HTML]{000000} 2.2986 & {\cellcolor[HTML]{79ABD0}} \color[HTML]{000000} 1.9514 & {\cellcolor[HTML]{3790C0}} \color[HTML]{000000} 2.4861 & {\cellcolor[HTML]{60A1CA}} \color[HTML]{000000} 2.1597 & {\cellcolor[HTML]{6BA5CD}} \color[HTML]{000000} 2.0694 & {\cellcolor[HTML]{62A2CB}} \color[HTML]{000000} 2.1528 & {\cellcolor[HTML]{5A9EC9}} \color[HTML]{000000} 2.2111 \\
Llama-3.1-70B-Instruct & {\cellcolor[HTML]{F4EDF6}} \color[HTML]{000000} 0.2986 & {\cellcolor[HTML]{DDDBEC}} \color[HTML]{000000} 0.7708 & {\cellcolor[HTML]{D0D1E6}} \color[HTML]{000000} 1.0000 & {\cellcolor[HTML]{F6EFF7}} \color[HTML]{000000} 0.2431 & {\cellcolor[HTML]{E9E5F1}} \color[HTML]{000000} 0.5556 & {\cellcolor[HTML]{E7E3F0}} \color[HTML]{000000} 0.5972 & {\cellcolor[HTML]{E6E2EF}} \color[HTML]{000000} 0.6181 & {\cellcolor[HTML]{EDE8F3}} \color[HTML]{000000} 0.4792 & {\cellcolor[HTML]{F0EAF4}} \color[HTML]{000000} 0.4028 & {\cellcolor[HTML]{FCF4FA}} \color[HTML]{000000} 0.0833 & {\cellcolor[HTML]{ECE7F2}} \color[HTML]{000000} 0.5049 \\
Llama-3.1-8B-Instruct & {\cellcolor[HTML]{FFF7FB}} \color[HTML]{000000} 0.0069 & {\cellcolor[HTML]{FCF4FA}} \color[HTML]{000000} 0.0833 & {\cellcolor[HTML]{D0D1E6}} \color[HTML]{000000} 1.0000 & {\cellcolor[HTML]{FFF7FB}} \color[HTML]{000000} 0.0000 & {\cellcolor[HTML]{FEF6FA}} \color[HTML]{000000} 0.0347 & {\cellcolor[HTML]{FEF6FB}} \color[HTML]{000000} 0.0208 & {\cellcolor[HTML]{FEF6FA}} \color[HTML]{000000} 0.0347 & {\cellcolor[HTML]{FFF7FB}} \color[HTML]{000000} 0.0000 & {\cellcolor[HTML]{FFF7FB}} \color[HTML]{000000} 0.0000 & {\cellcolor[HTML]{FCF4FA}} \color[HTML]{000000} 0.0833 & {\cellcolor[HTML]{FAF3F9}} \color[HTML]{000000} 0.1264 \\
Llama-3.2-1B-Instruct & {\cellcolor[HTML]{FFF7FB}} \color[HTML]{000000} 0.0000 & {\cellcolor[HTML]{FFF7FB}} \color[HTML]{000000} 0.0000 & {\cellcolor[HTML]{FEF6FA}} \color[HTML]{000000} 0.0417 & {\cellcolor[HTML]{FFF7FB}} \color[HTML]{000000} 0.0069 & {\cellcolor[HTML]{FFF7FB}} \color[HTML]{000000} 0.0139 & {\cellcolor[HTML]{FFF7FB}} \color[HTML]{000000} 0.0000 & {\cellcolor[HTML]{FFF7FB}} \color[HTML]{000000} 0.0000 & {\cellcolor[HTML]{FFF7FB}} \color[HTML]{000000} 0.0000 & {\cellcolor[HTML]{FFF7FB}} \color[HTML]{000000} 0.0000 & {\cellcolor[HTML]{FFF7FB}} \color[HTML]{000000} 0.0000 & {\cellcolor[HTML]{FFF7FB}} \color[HTML]{000000} 0.0062 \\
Llama-3.2-3B-Instruct & {\cellcolor[HTML]{FFF7FB}} \color[HTML]{000000} 0.0000 & {\cellcolor[HTML]{FBF4F9}} \color[HTML]{000000} 0.1042 & {\cellcolor[HTML]{DDDBEC}} \color[HTML]{000000} 0.7708 & {\cellcolor[HTML]{FEF6FB}} \color[HTML]{000000} 0.0208 & {\cellcolor[HTML]{FEF6FA}} \color[HTML]{000000} 0.0347 & {\cellcolor[HTML]{FFF7FB}} \color[HTML]{000000} 0.0000 & {\cellcolor[HTML]{FFF7FB}} \color[HTML]{000000} 0.0000 & {\cellcolor[HTML]{FFF7FB}} \color[HTML]{000000} 0.0139 & {\cellcolor[HTML]{FFF7FB}} \color[HTML]{000000} 0.0069 & {\cellcolor[HTML]{FDF5FA}} \color[HTML]{000000} 0.0556 & {\cellcolor[HTML]{FBF4F9}} \color[HTML]{000000} 0.1007 \\
Llama-3.3-70B-Instruct & {\cellcolor[HTML]{D3D4E7}} \color[HTML]{000000} 0.9444 & {\cellcolor[HTML]{AFC1DD}} \color[HTML]{000000} 1.3958 & {\cellcolor[HTML]{97B7D7}} \color[HTML]{000000} 1.6458 & {\cellcolor[HTML]{D7D6E9}} \color[HTML]{000000} 0.8819 & {\cellcolor[HTML]{CCCFE5}} \color[HTML]{000000} 1.0556 & {\cellcolor[HTML]{CCCFE5}} \color[HTML]{000000} 1.0556 & {\cellcolor[HTML]{CDD0E5}} \color[HTML]{000000} 1.0347 & {\cellcolor[HTML]{D2D3E7}} \color[HTML]{000000} 0.9583 & {\cellcolor[HTML]{D3D4E7}} \color[HTML]{000000} 0.9444 & {\cellcolor[HTML]{EDE8F3}} \color[HTML]{000000} 0.4722 & {\cellcolor[HTML]{CDD0E5}} \color[HTML]{000000} 1.0389 \\
Mistral-Nemo-Instruct-2407 & {\cellcolor[HTML]{EDE7F2}} \color[HTML]{000000} 0.4931 & {\cellcolor[HTML]{94B6D7}} \color[HTML]{000000} 1.6736 & {\cellcolor[HTML]{4C99C5}} \color[HTML]{000000} 2.3125 & {\cellcolor[HTML]{F3EDF5}} \color[HTML]{000000} 0.3194 & {\cellcolor[HTML]{EEE9F3}} \color[HTML]{000000} 0.4375 & {\cellcolor[HTML]{F0EAF4}} \color[HTML]{000000} 0.4028 & {\cellcolor[HTML]{E8E4F0}} \color[HTML]{000000} 0.5625 & {\cellcolor[HTML]{E4E1EF}} \color[HTML]{000000} 0.6458 & {\cellcolor[HTML]{EDE8F3}} \color[HTML]{000000} 0.4722 & {\cellcolor[HTML]{E4E1EF}} \color[HTML]{000000} 0.6458 & {\cellcolor[HTML]{DCDAEB}} \color[HTML]{000000} 0.7965 \\
Qwen3-1.7B & {\cellcolor[HTML]{F8F1F8}} \color[HTML]{000000} 0.1806 & {\cellcolor[HTML]{D9D8EA}} \color[HTML]{000000} 0.8542 & {\cellcolor[HTML]{73A9CF}} \color[HTML]{000000} 2.0069 & {\cellcolor[HTML]{EAE6F1}} \color[HTML]{000000} 0.5417 & {\cellcolor[HTML]{F4EEF6}} \color[HTML]{000000} 0.2847 & {\cellcolor[HTML]{F6EFF7}} \color[HTML]{000000} 0.2431 & {\cellcolor[HTML]{F5EFF6}} \color[HTML]{000000} 0.2639 & {\cellcolor[HTML]{ECE7F2}} \color[HTML]{000000} 0.5139 & {\cellcolor[HTML]{F1EBF5}} \color[HTML]{000000} 0.3681 & {\cellcolor[HTML]{FBF4F9}} \color[HTML]{000000} 0.0972 & {\cellcolor[HTML]{EAE6F1}} \color[HTML]{000000} 0.5354 \\
Qwen3-32B & {\cellcolor[HTML]{B9C6E0}} \color[HTML]{000000} 1.2778 & {\cellcolor[HTML]{94B6D7}} \color[HTML]{000000} 1.6806 & {\cellcolor[HTML]{6BA5CD}} \color[HTML]{000000} 2.0625 & {\cellcolor[HTML]{D2D2E7}} \color[HTML]{000000} 0.9722 & {\cellcolor[HTML]{B5C4DF}} \color[HTML]{000000} 1.3125 & {\cellcolor[HTML]{B8C6E0}} \color[HTML]{000000} 1.2917 & {\cellcolor[HTML]{ACC0DD}} \color[HTML]{000000} 1.4236 & {\cellcolor[HTML]{C5CCE3}} \color[HTML]{000000} 1.1250 & {\cellcolor[HTML]{B8C6E0}} \color[HTML]{000000} 1.2917 & {\cellcolor[HTML]{AFC1DD}} \color[HTML]{000000} 1.4028 & {\cellcolor[HTML]{B0C2DE}} \color[HTML]{000000} 1.3840 \\
Qwen3-4B & {\cellcolor[HTML]{E9E5F1}} \color[HTML]{000000} 0.5486 & {\cellcolor[HTML]{D5D5E8}} \color[HTML]{000000} 0.9167 & {\cellcolor[HTML]{76AAD0}} \color[HTML]{000000} 1.9792 & {\cellcolor[HTML]{F4EEF6}} \color[HTML]{000000} 0.2917 & {\cellcolor[HTML]{DAD9EA}} \color[HTML]{000000} 0.8264 & {\cellcolor[HTML]{EEE9F3}} \color[HTML]{000000} 0.4444 & {\cellcolor[HTML]{E1DFED}} \color[HTML]{000000} 0.6875 & {\cellcolor[HTML]{E7E3F0}} \color[HTML]{000000} 0.5903 & {\cellcolor[HTML]{EAE6F1}} \color[HTML]{000000} 0.5347 & {\cellcolor[HTML]{E4E1EF}} \color[HTML]{000000} 0.6458 & {\cellcolor[HTML]{DFDDEC}} \color[HTML]{000000} 0.7465 \\
\hline
\bfseries $\downarrow$ Average & {\cellcolor[HTML]{E0DDED}} \color[HTML]{000000} 0.7274 & {\cellcolor[HTML]{D1D2E6}} \color[HTML]{000000} 0.9896 & {\cellcolor[HTML]{A4BCDA}} \color[HTML]{000000} 1.5226 & {\cellcolor[HTML]{E7E3F0}} \color[HTML]{000000} 0.5964 & {\cellcolor[HTML]{DDDBEC}} \color[HTML]{000000} 0.7682 & {\cellcolor[HTML]{E2DFEE}} \color[HTML]{000000} 0.6871 & {\cellcolor[HTML]{DCDAEB}} \color[HTML]{000000} 0.7834 & {\cellcolor[HTML]{DDDBEC}} \color[HTML]{000000} 0.7656 & {\cellcolor[HTML]{E0DEED}} \color[HTML]{000000} 0.7057 & {\cellcolor[HTML]{E3E0EE}} \color[HTML]{000000} 0.6688 & {\cellcolor[HTML]{DAD9EA}} \color[HTML]{000000} 0.8215 \\
\hline
\end{tabular}
}
\caption{Per-language per-generator comparison of mean personalization capabilities (mean scores averaged across the two personalization types). Darker color gradient represents higher mean personalization quality.}
\label{tab:perlanguagegenerator}
\end{table*}

\begin{figure}[!t]
\centering
\includegraphics[width=\linewidth]{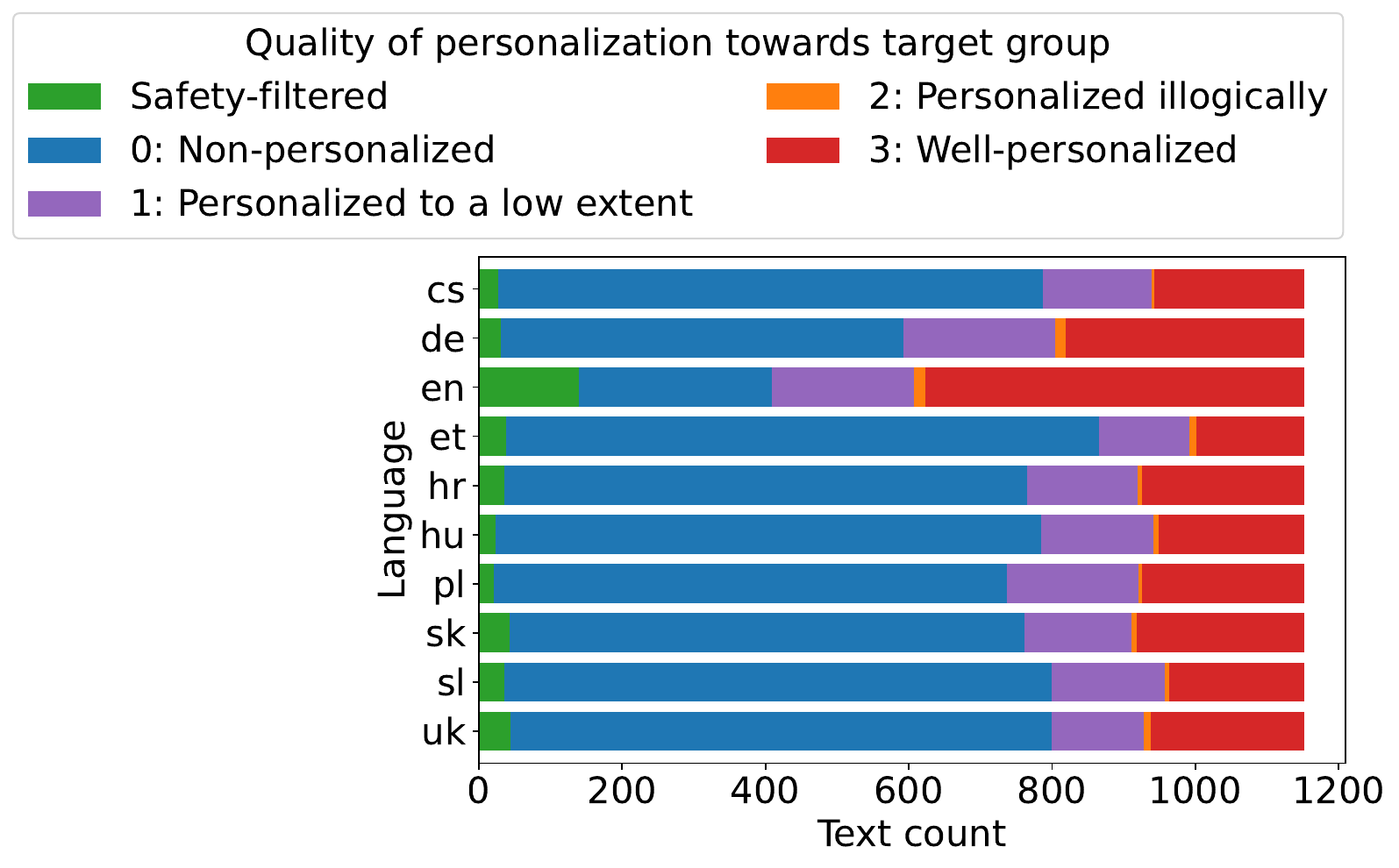}
\includegraphics[width=\linewidth]{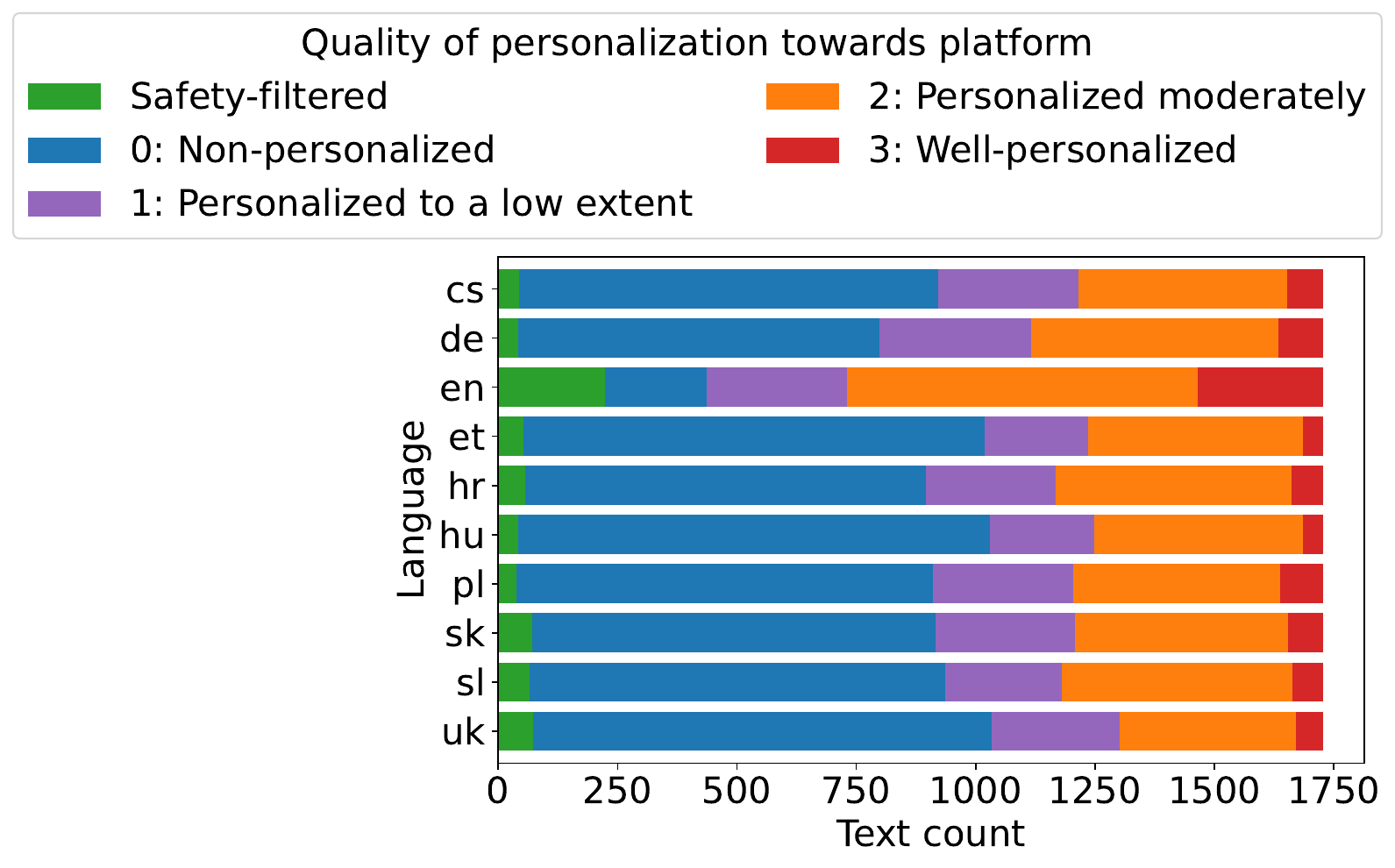}
\caption{Per-language comparison of personalization capabilities targeting groups (top, not including texts without target-group personalization request) and platforms (bottom).}
\label{fig:perlanguage}
\end{figure}

Analogously to the previous subsection, the evaluation of personalization capabilities across languages is illustrated in Fig.~\ref{fig:perlanguage}. In both personalization types, English (en) achieves the highest quality of personalization, followed by German (de). Both cases represent high-resource languages, dominating in the LLMs pretraining. On the other hand, Estonian as a low-resource language achieved the lowest quality of personalization.

In Table~\ref{tab:perlanguagegenerator}, personalization quality per each language for each LLM generator is provided. The values represent mean scores (values between 0 to 3, higher is better quality) of each generator averaged between the two personalization types. This comparison shows that the Gemma-3 models dominates in personalization capabilities across all tested languages. All the used LLMs have better personalization capabilities in English than the other languages.

\textbf{There are significant differences in personalization capabilities between English and non-English languages.} Statistical significance was tested by a paired t-test with $\alpha$ = 0.05. Although the results might be affected by metaevaluation of personalization quality (also performing possibly better for English), the differences are clear by a large margin. All the models have consistently better personalization capabilities in English. For example, the Mistral model achieved almost the same personalization quality as the Gemma-3 models for English, but significantly lower for German, and much lower for other languages.

\subsection{Detectability Analysis (RQ3)}

\begin{table*}[!t]
\centering
\resizebox{\textwidth}{!}{
\addtolength{\tabcolsep}{-2pt}
\begin{tabular}{l||c|c|c|c|c|c|c|c|c|c||c}
\hline
& \multicolumn{10}{c||}{\textbf{Language}}\\
\textbf{Generator} & \textbf{cs} & \textbf{de} & \textbf{en} & \textbf{et} & \textbf{hr} & \textbf{hu} & \textbf{pl} & \textbf{sk} & \textbf{sl} & \textbf{uk} & \textbf{$\rightarrow$ Average} \\
\hline
DeepSeek-R1-Distill-Llama-8B & {\cellcolor[HTML]{A4BCDA}} \color[HTML]{000000} 0.7593 & {\cellcolor[HTML]{94B6D7}} \color[HTML]{000000} 0.8025 & {\cellcolor[HTML]{A4BCDA}} \color[HTML]{000000} 0.7562 & {\cellcolor[HTML]{A4BCDA}} \color[HTML]{000000} 0.7562 & {\cellcolor[HTML]{C1CAE2}} \color[HTML]{000000} 0.6543 & {\cellcolor[HTML]{88B1D4}} \color[HTML]{000000} 0.8395 & {\cellcolor[HTML]{84B0D3}} \color[HTML]{000000} 0.8488 & {\cellcolor[HTML]{A4BCDA}} \color[HTML]{000000} 0.7593 & {\cellcolor[HTML]{C9CEE4}} \color[HTML]{000000} 0.6265 & {\cellcolor[HTML]{B9C6E0}} \color[HTML]{000000} 0.6821 & {\cellcolor[HTML]{A7BDDB}} \color[HTML]{000000} 0.7485 \\
DeepSeek-R1-Distill-Qwen-32B & {\cellcolor[HTML]{8EB3D5}} \color[HTML]{000000} 0.8241 & {\cellcolor[HTML]{A7BDDB}} \color[HTML]{000000} 0.7469 & {\cellcolor[HTML]{C0C9E2}} \color[HTML]{000000} 0.6605 & {\cellcolor[HTML]{94B6D7}} \color[HTML]{000000} 0.8025 & {\cellcolor[HTML]{84B0D3}} \color[HTML]{000000} 0.8488 & {\cellcolor[HTML]{76AAD0}} \color[HTML]{000000} 0.8920 & {\cellcolor[HTML]{B8C6E0}} \color[HTML]{000000} 0.6883 & {\cellcolor[HTML]{84B0D3}} \color[HTML]{000000} 0.8519 & {\cellcolor[HTML]{7DACD1}} \color[HTML]{000000} 0.8735 & {\cellcolor[HTML]{B0C2DE}} \color[HTML]{000000} 0.7160 & {\cellcolor[HTML]{99B8D8}} \color[HTML]{000000} 0.7904 \\
Gemma-2-27B-it & {\cellcolor[HTML]{67A4CC}} \color[HTML]{000000} 0.9321 & {\cellcolor[HTML]{81AED2}} \color[HTML]{000000} 0.8580 & {\cellcolor[HTML]{78ABD0}} \color[HTML]{000000} 0.8889 & {\cellcolor[HTML]{7DACD1}} \color[HTML]{000000} 0.8765 & {\cellcolor[HTML]{69A5CC}} \color[HTML]{000000} 0.9259 & {\cellcolor[HTML]{71A8CE}} \color[HTML]{000000} 0.9074 & {\cellcolor[HTML]{63A2CB}} \color[HTML]{000000} 0.9383 & {\cellcolor[HTML]{76AAD0}} \color[HTML]{000000} 0.8951 & {\cellcolor[HTML]{73A9CF}} \color[HTML]{000000} 0.9012 & {\cellcolor[HTML]{6FA7CE}} \color[HTML]{000000} 0.9105 & {\cellcolor[HTML]{73A9CF}} \color[HTML]{000000} 0.9034 \\
Gemma-2-2B-it & {\cellcolor[HTML]{8FB4D6}} \color[HTML]{000000} 0.8179 & {\cellcolor[HTML]{9EBAD9}} \color[HTML]{000000} 0.7747 & {\cellcolor[HTML]{ADC1DD}} \color[HTML]{000000} 0.7222 & {\cellcolor[HTML]{83AFD3}} \color[HTML]{000000} 0.8549 & {\cellcolor[HTML]{94B6D7}} \color[HTML]{000000} 0.8056 & {\cellcolor[HTML]{8BB2D4}} \color[HTML]{000000} 0.8333 & {\cellcolor[HTML]{91B5D6}} \color[HTML]{000000} 0.8117 & {\cellcolor[HTML]{8BB2D4}} \color[HTML]{000000} 0.8333 & {\cellcolor[HTML]{99B8D8}} \color[HTML]{000000} 0.7901 & {\cellcolor[HTML]{9EBAD9}} \color[HTML]{000000} 0.7778 & {\cellcolor[HTML]{94B6D7}} \color[HTML]{000000} 0.8022 \\
Gemma-2-9B-it & {\cellcolor[HTML]{B3C3DE}} \color[HTML]{000000} 0.7068 & {\cellcolor[HTML]{D9D8EA}} \color[HTML]{000000} 0.5556 & {\cellcolor[HTML]{CED0E6}} \color[HTML]{000000} 0.6080 & {\cellcolor[HTML]{97B7D7}} \color[HTML]{000000} 0.7963 & {\cellcolor[HTML]{B0C2DE}} \color[HTML]{000000} 0.7160 & {\cellcolor[HTML]{C0C9E2}} \color[HTML]{000000} 0.6605 & {\cellcolor[HTML]{BDC8E1}} \color[HTML]{000000} 0.6667 & {\cellcolor[HTML]{BDC8E1}} \color[HTML]{000000} 0.6667 & {\cellcolor[HTML]{C1CAE2}} \color[HTML]{000000} 0.6543 & {\cellcolor[HTML]{C9CEE4}} \color[HTML]{000000} 0.6235 & {\cellcolor[HTML]{BFC9E1}} \color[HTML]{000000} 0.6654 \\
Gemma-3-27B-it & {\cellcolor[HTML]{C5CCE3}} \color[HTML]{000000} 0.6389 & {\cellcolor[HTML]{F5EFF6}} \color[HTML]{000000} 0.3765 & {\cellcolor[HTML]{F6EFF7}} \color[HTML]{000000} 0.3704 & {\cellcolor[HTML]{97B7D7}} \color[HTML]{000000} 0.7932 & {\cellcolor[HTML]{D4D4E8}} \color[HTML]{000000} 0.5802 & {\cellcolor[HTML]{B1C2DE}} \color[HTML]{000000} 0.7099 & {\cellcolor[HTML]{D6D6E9}} \color[HTML]{000000} 0.5679 & {\cellcolor[HTML]{D1D2E6}} \color[HTML]{000000} 0.5957 & {\cellcolor[HTML]{EAE6F1}} \color[HTML]{000000} 0.4630 & {\cellcolor[HTML]{EAE6F1}} \color[HTML]{000000} 0.4599 & {\cellcolor[HTML]{D9D8EA}} \color[HTML]{000000} 0.5556 \\
Gemma-3-4B-it & {\cellcolor[HTML]{B9C6E0}} \color[HTML]{000000} 0.6821 & {\cellcolor[HTML]{CED0E6}} \color[HTML]{000000} 0.6049 & {\cellcolor[HTML]{C9CEE4}} \color[HTML]{000000} 0.6235 & {\cellcolor[HTML]{8BB2D4}} \color[HTML]{000000} 0.8302 & {\cellcolor[HTML]{B5C4DF}} \color[HTML]{000000} 0.6975 & {\cellcolor[HTML]{8BB2D4}} \color[HTML]{000000} 0.8333 & {\cellcolor[HTML]{D0D1E6}} \color[HTML]{000000} 0.6019 & {\cellcolor[HTML]{C9CEE4}} \color[HTML]{000000} 0.6265 & {\cellcolor[HTML]{B7C5DF}} \color[HTML]{000000} 0.6914 & {\cellcolor[HTML]{CED0E6}} \color[HTML]{000000} 0.6080 & {\cellcolor[HTML]{B9C6E0}} \color[HTML]{000000} 0.6799 \\
Llama-3.1-70B-Instruct & {\cellcolor[HTML]{89B1D4}} \color[HTML]{000000} 0.8364 & {\cellcolor[HTML]{9EBAD9}} \color[HTML]{000000} 0.7747 & {\cellcolor[HTML]{7BACD1}} \color[HTML]{000000} 0.8796 & {\cellcolor[HTML]{9CB9D9}} \color[HTML]{000000} 0.7809 & {\cellcolor[HTML]{91B5D6}} \color[HTML]{000000} 0.8148 & {\cellcolor[HTML]{84B0D3}} \color[HTML]{000000} 0.8488 & {\cellcolor[HTML]{81AED2}} \color[HTML]{000000} 0.8580 & {\cellcolor[HTML]{7DACD1}} \color[HTML]{000000} 0.8735 & {\cellcolor[HTML]{8FB4D6}} \color[HTML]{000000} 0.8179 & {\cellcolor[HTML]{A5BDDB}} \color[HTML]{000000} 0.7500 & {\cellcolor[HTML]{8EB3D5}} \color[HTML]{000000} 0.8235 \\
Llama-3.1-8B-Instruct & {\cellcolor[HTML]{C2CBE2}} \color[HTML]{000000} 0.6481 & {\cellcolor[HTML]{A4BCDA}} \color[HTML]{000000} 0.7562 & {\cellcolor[HTML]{62A2CB}} \color[HTML]{000000} 0.9444 & {\cellcolor[HTML]{E0DDED}} \color[HTML]{000000} 0.5185 & {\cellcolor[HTML]{B7C5DF}} \color[HTML]{000000} 0.6914 & {\cellcolor[HTML]{B7C5DF}} \color[HTML]{000000} 0.6914 & {\cellcolor[HTML]{A9BFDC}} \color[HTML]{000000} 0.7377 & {\cellcolor[HTML]{CACEE5}} \color[HTML]{000000} 0.6204 & {\cellcolor[HTML]{BDC8E1}} \color[HTML]{000000} 0.6698 & {\cellcolor[HTML]{B3C3DE}} \color[HTML]{000000} 0.7068 & {\cellcolor[HTML]{B4C4DF}} \color[HTML]{000000} 0.6985 \\
Llama-3.2-1B-Instruct & {\cellcolor[HTML]{DFDDEC}} \color[HTML]{000000} 0.5247 & {\cellcolor[HTML]{94B6D7}} \color[HTML]{000000} 0.8025 & {\cellcolor[HTML]{80AED2}} \color[HTML]{000000} 0.8642 & {\cellcolor[HTML]{CCCFE5}} \color[HTML]{000000} 0.6142 & {\cellcolor[HTML]{C9CEE4}} \color[HTML]{000000} 0.6235 & {\cellcolor[HTML]{CED0E6}} \color[HTML]{000000} 0.6080 & {\cellcolor[HTML]{BBC7E0}} \color[HTML]{000000} 0.6759 & {\cellcolor[HTML]{E3E0EE}} \color[HTML]{000000} 0.5000 & {\cellcolor[HTML]{D5D5E8}} \color[HTML]{000000} 0.5741 & {\cellcolor[HTML]{E1DFED}} \color[HTML]{000000} 0.5093 & {\cellcolor[HTML]{C8CDE4}} \color[HTML]{000000} 0.6296 \\
Llama-3.2-3B-Instruct & {\cellcolor[HTML]{DFDDEC}} \color[HTML]{000000} 0.5247 & {\cellcolor[HTML]{8BB2D4}} \color[HTML]{000000} 0.8333 & {\cellcolor[HTML]{67A4CC}} \color[HTML]{000000} 0.9290 & {\cellcolor[HTML]{DFDDEC}} \color[HTML]{000000} 0.5216 & {\cellcolor[HTML]{D2D2E7}} \color[HTML]{000000} 0.5926 & {\cellcolor[HTML]{CCCFE5}} \color[HTML]{000000} 0.6173 & {\cellcolor[HTML]{E0DEED}} \color[HTML]{000000} 0.5123 & {\cellcolor[HTML]{D8D7E9}} \color[HTML]{000000} 0.5617 & {\cellcolor[HTML]{BBC7E0}} \color[HTML]{000000} 0.6790 & {\cellcolor[HTML]{E5E1EF}} \color[HTML]{000000} 0.4907 & {\cellcolor[HTML]{C9CEE4}} \color[HTML]{000000} 0.6262 \\
Llama-3.3-70B-Instruct & {\cellcolor[HTML]{67A4CC}} \color[HTML]{000000} 0.9290 & {\cellcolor[HTML]{6BA5CD}} \color[HTML]{000000} 0.9198 & {\cellcolor[HTML]{67A4CC}} \color[HTML]{000000} 0.9321 & {\cellcolor[HTML]{78ABD0}} \color[HTML]{000000} 0.8889 & {\cellcolor[HTML]{7EADD1}} \color[HTML]{000000} 0.8704 & {\cellcolor[HTML]{589EC8}} \color[HTML]{000000} 0.9691 & {\cellcolor[HTML]{5C9FC9}} \color[HTML]{000000} 0.9599 & {\cellcolor[HTML]{6BA5CD}} \color[HTML]{000000} 0.9228 & {\cellcolor[HTML]{8CB3D5}} \color[HTML]{000000} 0.8272 & {\cellcolor[HTML]{9AB8D8}} \color[HTML]{000000} 0.7840 & {\cellcolor[HTML]{73A9CF}} \color[HTML]{000000} 0.9003 \\
Mistral-Nemo-Instruct-2407 & {\cellcolor[HTML]{81AED2}} \color[HTML]{000000} 0.8611 & {\cellcolor[HTML]{9FBAD9}} \color[HTML]{000000} 0.7716 & {\cellcolor[HTML]{C0C9E2}} \color[HTML]{000000} 0.6574 & {\cellcolor[HTML]{8EB3D5}} \color[HTML]{000000} 0.8241 & {\cellcolor[HTML]{A1BBDA}} \color[HTML]{000000} 0.7685 & {\cellcolor[HTML]{88B1D4}} \color[HTML]{000000} 0.8426 & {\cellcolor[HTML]{9CB9D9}} \color[HTML]{000000} 0.7809 & {\cellcolor[HTML]{91B5D6}} \color[HTML]{000000} 0.8148 & {\cellcolor[HTML]{9EBAD9}} \color[HTML]{000000} 0.7747 & {\cellcolor[HTML]{88B1D4}} \color[HTML]{000000} 0.8395 & {\cellcolor[HTML]{97B7D7}} \color[HTML]{000000} 0.7935 \\
Qwen3-1.7B & {\cellcolor[HTML]{549CC7}} \color[HTML]{000000} 0.9753 & {\cellcolor[HTML]{5EA0CA}} \color[HTML]{000000} 0.9537 & {\cellcolor[HTML]{7DACD1}} \color[HTML]{000000} 0.8765 & {\cellcolor[HTML]{6FA7CE}} \color[HTML]{000000} 0.9105 & {\cellcolor[HTML]{549CC7}} \color[HTML]{000000} 0.9784 & {\cellcolor[HTML]{509AC6}} \color[HTML]{000000} 0.9877 & {\cellcolor[HTML]{549CC7}} \color[HTML]{000000} 0.9753 & {\cellcolor[HTML]{549CC7}} \color[HTML]{000000} 0.9753 & {\cellcolor[HTML]{60A1CA}} \color[HTML]{000000} 0.9475 & {\cellcolor[HTML]{4E9AC6}} \color[HTML]{000000} 0.9907 & {\cellcolor[HTML]{5C9FC9}} \color[HTML]{000000} 0.9571 \\
Qwen3-32B & {\cellcolor[HTML]{97B7D7}} \color[HTML]{000000} 0.7932 & {\cellcolor[HTML]{D3D4E7}} \color[HTML]{000000} 0.5833 & {\cellcolor[HTML]{E8E4F0}} \color[HTML]{000000} 0.4691 & {\cellcolor[HTML]{79ABD0}} \color[HTML]{000000} 0.8827 & {\cellcolor[HTML]{9AB8D8}} \color[HTML]{000000} 0.7840 & {\cellcolor[HTML]{80AED2}} \color[HTML]{000000} 0.8642 & {\cellcolor[HTML]{9EBAD9}} \color[HTML]{000000} 0.7747 & {\cellcolor[HTML]{91B5D6}} \color[HTML]{000000} 0.8117 & {\cellcolor[HTML]{A5BDDB}} \color[HTML]{000000} 0.7531 & {\cellcolor[HTML]{AFC1DD}} \color[HTML]{000000} 0.7191 & {\cellcolor[HTML]{A8BEDC}} \color[HTML]{000000} 0.7435 \\
Qwen3-4B & {\cellcolor[HTML]{589EC8}} \color[HTML]{000000} 0.9691 & {\cellcolor[HTML]{76AAD0}} \color[HTML]{000000} 0.8920 & {\cellcolor[HTML]{A5BDDB}} \color[HTML]{000000} 0.7531 & {\cellcolor[HTML]{4C99C5}} \color[HTML]{000000} 0.9938 & {\cellcolor[HTML]{67A4CC}} \color[HTML]{000000} 0.9321 & {\cellcolor[HTML]{4C99C5}} \color[HTML]{000000} 0.9938 & {\cellcolor[HTML]{63A2CB}} \color[HTML]{000000} 0.9414 & {\cellcolor[HTML]{509AC6}} \color[HTML]{000000} 0.9877 & {\cellcolor[HTML]{589EC8}} \color[HTML]{000000} 0.9660 & {\cellcolor[HTML]{569DC8}} \color[HTML]{000000} 0.9722 & {\cellcolor[HTML]{63A2CB}} \color[HTML]{000000} 0.9401 \\
\hline
\bfseries $\downarrow$ Average & {\cellcolor[HTML]{9EBAD9}} \color[HTML]{000000} 0.7764 & {\cellcolor[HTML]{A5BDDB}} \color[HTML]{000000} 0.7504 & {\cellcolor[HTML]{A7BDDB}} \color[HTML]{000000} 0.7459 & {\cellcolor[HTML]{99B8D8}} \color[HTML]{000000} 0.7903 & {\cellcolor[HTML]{A1BBDA}} \color[HTML]{000000} 0.7677 & {\cellcolor[HTML]{8FB4D6}} \color[HTML]{000000} 0.8187 & {\cellcolor[HTML]{9FBAD9}} \color[HTML]{000000} 0.7712 & {\cellcolor[HTML]{A1BBDA}} \color[HTML]{000000} 0.7685 & {\cellcolor[HTML]{A5BDDB}} \color[HTML]{000000} 0.7506 & {\cellcolor[HTML]{AFC1DD}} \color[HTML]{000000} 0.7213 & {\cellcolor[HTML]{A1BBDA}} \color[HTML]{000000} 0.7661 \\
\hline
\end{tabular}
}
\caption{Per-language per-generator evaluation of generated-texts detectability (average TPR across the three detectors). Darker color gradient represents higher TPR.}
\label{tab:detectability_generators_languages}
\end{table*}

To evaluate the effect of personalization on detectability of machine-generated texts, we averaged the predictions by three existing multilingual detector for each text. The results are represented in the form of true positive rate, i.e., a proportion of the texts that are detected as machine-generated by existing detectors. The results summarized in Table~\ref{tab:detectability_generators_languages} indicate that about $3/4$ of the texts are correctly detected.

\textbf{The detectability of generated personalized texts varies across languages.}
On average, the texts in Ukrainian (uk) are the most difficult to detect (TPR of 0.72), followed by English, German, and Slovenian (TPR of 0.75). The best detectability was observed in case of Hungarian texts (TPR of 0.82). However, the results are not consistent across the LLM generators. The most difficult do detect are the texts generated by the Gemma-3-27B-it model, but also the smaller Llama-3.2 models. The easiest to detect are the text generated by the smaller Qwen3 models. Therefore, the detectability is not affected by personalization only (the best quality of Gemma-3 vs the lowest quality of Llama-3.2 models), but also the presence of noise and failed translations.

The actual effect of personalization (both types, targeting groups and platforms) is illustrated in Table~\ref{tab:detectability_groups_platforms}. The results indicate that Eurpean conservatives target group personalization makes the texts the easiest to detect, while targeting Twitter makes the texts consistently the most difficult to detect. The Twitter platform has a stricter length-limitation of the texts, which naturally makes the detection challenging (shorter texts). Interestingly, the European conservatives target group achieve the highest group-personalization quality, while the worst platform-personalization quality. It also has the lowest linguistic-acceptability and output-content-quality scores, as well as the highest amount of noise.

\textbf{Personalization-quality towards platforms influences detectability more than personalization-quality towards target groups.} When removing noisy and safety-filtered texts, we observe a consistent and significant decrease of true positive rate for the increasing platform-personalization quality (TPR averaging from 0.93 for 0-score quality to 0.67 for 3-score quality). We do not observe such a clear correlation for group-personalization quality, although the well-personalized texts have worse detectability than the non-personalized texts (average TPR of 0.72 vs 0.8, respectively).

\begin{table}[!t]
\centering
\resizebox{\linewidth}{!}{
\addtolength{\tabcolsep}{-2pt}
\begin{tabular}{l||c|c|c||c}
\hline
& \multicolumn{3}{c||}{\textbf{Target platform}}\\
\textbf{Target group} & \textbf{Mastodon} & \textbf{Telegram} & \textbf{Twitter} & \textbf{$\rightarrow$ Average} \\
\hline
None & \bfseries 0.7677 & 0.7665 & \bfseries 0.7288 & \bfseries 0.7543 \\
European conservatives & 0.7917 & 0.7922 & 0.7708 & 0.7849 \\
Urban population & 0.7710 & \bfseries 0.7658 & 0.7405 & 0.7591 \\
\hline
\bfseries $\downarrow$ Average & 0.7768 & 0.7748 & 0.7467 & 0.7661 \\
\hline
\end{tabular}
}
\caption{Per-platform per-group evaluation of generated-texts detectability (average TPR across the three detectors). Bold represents the lowest value per each column.}
\label{tab:detectability_groups_platforms}
\end{table}

The results in Table~\ref{tab:detectability_stance_narratives} also show differences in detectability based on stance towards the disinformation narrative, where the texts supporting the narrative have better detectability than those against the narrative. This behavior can be considered safer than in other case. There are also some noticeable differences in detectability based on the narratives themselves (average TPR ranging from 0.74 to 0.8).

\begin{table}[!t]
\centering
\resizebox{\linewidth}{!}{
\addtolength{\tabcolsep}{-2pt}
\begin{tabular}{p{5cm}||c|c||c}
\hline
& \multicolumn{2}{c||}{\textbf{Stance}}\\
\textbf{Narrative} & \textbf{Against} & \textbf{Supporting} & \textbf{$\rightarrow$ Average} \\
\hline
Bucha massacre was staged & 0.7836 & 0.8257 & 0.8046 \\
Cannabis is a 'cancer killer' & 0.7292 & 0.7664 & 0.7478 \\
EU wants to conceal the presence of the insects in products... & 0.7387 & \bfseries 0.7331 & \bfseries 0.7359 \\
People die after being vaccinated against COVID-19 & 0.7748 & 0.8257 & 0.8002 \\
Planes are spraying chemtrails & \bfseries 0.7269 & 0.7831 & 0.7550 \\
Ukraine hosts secret US bio-labs & 0.7354 & 0.7708 & 0.7531 \\
\hline
\bfseries $\downarrow$ Average & 0.7481 & 0.7841 & 0.7661 \\
\hline
\end{tabular}
}
\caption{Per-stance per-narrative evaluation of generated-texts detectability (average TPR across the three detectors). Bold represents the lowest value per each column.}
\label{tab:detectability_stance_narratives}
\end{table}

\section{Conclusions}
\label{sec:conclusion}

This study, focused on multilingual personalization capabilities of text-generation LLMs, revealed differences in capabilities when personalizing the text towards specific target demographic groups and towards specific social-media platforms. The results indicate significantly higher personalization capabilities for texts generated in English than non-English languages. Also, the detectability of generated texts is lower for well-personalized text in English than the other languages. Ukrainian is however the most difficult to detect on average (maybe due to the Cyrillic script). Platform-based personalization decreases the detectability in higher amount than group-based personalization.

\clearpage
\section*{Limitations}
\label{sec:limitations}
Our key findings rely on an LLM-based metaevaluation of personalization quality. Although we have used 3 LLMs to limit the effect of internal biases in evaluation, there can still be some. The study is limited to 10 languages of 3 language-family branches; however, we are unsure about generalization of our findings to other languages. We have limited the generators to 16 SOTA LLMs. The results for other LLMs might differ.

\section*{Ethics Statement}
\label{sec:ethics}
To balance the negative effect of evaluation of personalization capabilities of LLM while generating disinformation, we evaluate the same also while generating counter-disinformation texts (usable for argumentation and intervention). The artifacts and results of this study are intended for research purpose only to evaluate personalization capabilities of existing LLMs. The existing artifacts used in this work have been properly cited and used according their licenses and intended use. We have also checked and followed licensing and terms of use of the used LLMs. AI assistants have not been used for conducting research in any other way than already described in the paper (text generation and metaevaluation).

\section*{Acknowledgments}
Funded by the EU NextGenerationEU through the Recovery and Resilience Plan for Slovakia under the project No. 09I01-03-V04-00068.

\textbf{Computational resources}. We acknowledge EuroHPC Joint Undertaking for awarding us access to Leonardo at CINECA, Italy. Part of the research results was obtained using the computational resources procured in the national project \textit{National competence centre for high performance computing} (project code: 311070AKF2) funded by European Regional Development Fund, EU Structural Funds Informatization of Society, Operational Program Integrated Infrastructure.

\bibliography{anthology,custom}

\begin{thebibliography}{30}
\providecommand{\natexlab}[1]{#1}

\bibitem[{mis(2024)}]{mistralnemo}
 2024.
\newblock \href {https://mistral.ai/news/mistral-nemo} {Mistral {NeMo}: our new best small model. a state-of-the-art {12B} model with 128k context length, built in collaboration with {NVIDIA}, and released under the {Apache} 2.0 license.}
\newblock Release blog.

\bibitem[{Bao et~al.(2024)Bao, Zhao, Teng, Yang, and Zhang}]{bao2024fastdetectgpt}
Guangsheng Bao, Yanbin Zhao, Zhiyang Teng, Linyi Yang, and Yue Zhang. 2024.
\newblock \href {https://openreview.net/forum?id=Bpcgcr8E8Z} {Fast-{D}etect{GPT}: Efficient zero-shot detection of machine-generated text via conditional probability curvature}.
\newblock In \emph{The Twelfth International Conference on Learning Representations}.

\bibitem[{Barman(2025)}]{10.1145/3726831.3726835}
Dipto Barman. 2025.
\newblock \href {https://doi.org/10.1145/3726831.3726835} {Rethinking misinformation mitigation: The case for personalized digital interventions}.
\newblock \emph{SIGWEB Newsl.}, 2025(Winter).

\bibitem[{Bideleux and Jeffries(2007)}]{bideleux2007history}
Robert Bideleux and Ian Jeffries. 2007.
\newblock \emph{A history of Eastern Europe: Crisis and change}.
\newblock Routledge.

\bibitem[{Blom(2000)}]{10.1145/633292.633483}
Jan Blom. 2000.
\newblock \href {https://doi.org/10.1145/633292.633483} {Personalization: a taxonomy}.
\newblock In \emph{CHI '00 Extended Abstracts on Human Factors in Computing Systems}, CHI EA '00, page 313–314, New York, NY, USA. Association for Computing Machinery.

\bibitem[{Cima et~al.(2025)Cima, Miaschi, Trujillo, Avvenuti, Dell'Orletta, and Cresci}]{10.1145/3696410.3714507}
Lorenzo Cima, Alessio Miaschi, Amaury Trujillo, Marco Avvenuti, Felice Dell'Orletta, and Stefano Cresci. 2025.
\newblock \href {https://doi.org/10.1145/3696410.3714507} {Contextualized counterspeech: Strategies for adaptation, personalization, and evaluation}.
\newblock In \emph{Proceedings of the ACM on Web Conference 2025}, WWW '25, page 5022–5033, New York, NY, USA. Association for Computing Machinery.

\bibitem[{Fu et~al.(2025)Fu, Rahmani, Wu, Ramos, Yilmaz, and Lipani}]{fu2025prefreferencefreeevaluationpersonalised}
Xiao Fu, Hossein~A. Rahmani, Bin Wu, Jerome Ramos, Emine Yilmaz, and Aldo Lipani. 2025.
\newblock \href {https://arxiv.org/abs/2508.10028} {{PREF}: Reference-free evaluation of personalised text generation in {LLMs}}.
\newblock \emph{Preprint}, arXiv:2508.10028.

\bibitem[{Gajewska et~al.(2025)Gajewska, Derbent, Chudziak, and Budzynska}]{gajewska2025algorithmicfairnessnlppersonainfused}
Ewelina Gajewska, Arda Derbent, Jaroslaw~A Chudziak, and Katarzyna Budzynska. 2025.
\newblock \href {https://arxiv.org/abs/2510.19331} {Algorithmic fairness in {NLP}: Persona-infused {LLMs} for human-centric hate speech detection}.
\newblock \emph{Preprint}, arXiv:2510.19331.

\bibitem[{Gao et~al.(2025)Gao, Li, Wang, Li, Liu, Song, Zhang, Yan, Nakov, and Chen}]{gao2025personalizationtricksdetectorsfeatureinversion}
Lang Gao, Xuhui Li, Chenxi Wang, Mingzhe Li, Wei Liu, Zirui Song, Jinghui Zhang, Rui Yan, Preslav Nakov, and Xiuying Chen. 2025.
\newblock \href {https://arxiv.org/abs/2510.12476} {When personalization tricks detectors: The feature-inversion trap in machine-generated text detection}.
\newblock \emph{Preprint}, arXiv:2510.12476.

\bibitem[{Grattafiori et~al.(2024)Grattafiori, Dubey, Jauhri, Pandey, Kadian, Al-Dahle, Letman, Mathur, Schelten, Vaughan, Yang, Fan, Goyal, Hartshorn, Yang, Mitra, Sravankumar, Korenev, Hinsvark, Rao, Zhang, Rodriguez, Gregerson, Spataru, Roziere, Biron, Tang, Chern, Caucheteux, Nayak, Bi, Marra, McConnell, Keller, Touret, Wu, Wong, Ferrer, Nikolaidis, Allonsius, Song, Pintz, Livshits, Wyatt, Esiobu, Choudhary, Mahajan, Garcia-Olano, Perino, Hupkes, Lakomkin, AlBadawy, Lobanova, Dinan, Smith, Radenovic, Guzmán, Zhang, Synnaeve, Lee, Anderson, Thattai, Nail, Mialon, Pang, Cucurell, Nguyen, Korevaar, Xu, Touvron, Zarov, Ibarra, Kloumann, Misra, Evtimov, Zhang, Copet, Lee, Geffert, Vranes, Park, Mahadeokar, Shah, van~der Linde, Billock, Hong, Lee, Fu, Chi, Huang, Liu, Wang, Yu, Bitton, Spisak, Park, Rocca, Johnstun, Saxe, Jia, Alwala, Prasad, Upasani, Plawiak, Li, Heafield, Stone, El-Arini, Iyer, Malik, Chiu, Bhalla, Lakhotia, Rantala-Yeary, van~der Maaten, Chen, Tan, Jenkins, Martin, Madaan, Malo, Blecher,
  Landzaat, de~Oliveira, Muzzi, Pasupuleti, Singh, Paluri, Kardas, Tsimpoukelli, Oldham, Rita, Pavlova, Kambadur, Lewis, Si, Singh, Hassan, Goyal, Torabi, Bashlykov, Bogoychev, Chatterji, Zhang, Duchenne, Çelebi, Alrassy, Zhang, Li, Vasic, Weng, Bhargava, Dubal, Krishnan, Koura, Xu, He, Dong, Srinivasan, Ganapathy, Calderer, Cabral, Stojnic, Raileanu, Maheswari, Girdhar, Patel, Sauvestre, Polidoro, Sumbaly, Taylor, Silva, Hou, Wang, Hosseini, Chennabasappa, Singh, Bell, Kim, Edunov, Nie, Narang, Raparthy, Shen, Wan, Bhosale, Zhang, Vandenhende, Batra, Whitman, Sootla, Collot, Gururangan, Borodinsky, Herman, Fowler, Sheasha, Georgiou, Scialom, Speckbacher, Mihaylov, Xiao, Karn, Goswami, Gupta, Ramanathan, Kerkez, Gonguet, Do, Vogeti, Albiero, Petrovic, Chu, Xiong, Fu, Meers, Martinet, Wang, Wang, Tan, Xia, Xie, Jia, Wang, Goldschlag, Gaur, Babaei, Wen, Song, Zhang, Li, Mao, Coudert, Yan, Chen, Papakipos, Singh, Srivastava, Jain, Kelsey, Shajnfeld, Gangidi, Victoria, Goldstand, Menon, Sharma, Boesenberg,
  Baevski, Feinstein, Kallet, Sangani, Teo, Yunus, Lupu, Alvarado, Caples, Gu, Ho, Poulton, Ryan, Ramchandani, Dong, Franco, Goyal, Saraf, Chowdhury, Gabriel, Bharambe, Eisenman, Yazdan, James, Maurer, Leonhardi, Huang, Loyd, Paola, Paranjape, Liu, Wu, Ni, Hancock, Wasti, Spence, Stojkovic, Gamido, Montalvo, Parker, Burton, Mejia, Liu, Wang, Kim, Zhou, Hu, Chu, Cai, Tindal, Feichtenhofer, Gao, Civin, Beaty, Kreymer, Li, Adkins, Xu, Testuggine, David, Parikh, Liskovich, Foss, Wang, Le, Holland, Dowling, Jamil, Montgomery, Presani, Hahn, Wood, Le, Brinkman, Arcaute, Dunbar, Smothers, Sun, Kreuk, Tian, Kokkinos, Ozgenel, Caggioni, Kanayet, Seide, Florez, Schwarz, Badeer, Swee, Halpern, Herman, Sizov, Guangyi, Zhang, Lakshminarayanan, Inan, Shojanazeri, Zou, Wang, Zha, Habeeb, Rudolph, Suk, Aspegren, Goldman, Zhan, Damlaj, Molybog, Tufanov, Leontiadis, Veliche, Gat, Weissman, Geboski, Kohli, Lam, Asher, Gaya, Marcus, Tang, Chan, Zhen, Reizenstein, Teboul, Zhong, Jin, Yang, Cummings, Carvill, Shepard, McPhie,
  Torres, Ginsburg, Wang, Wu, U, Saxena, Khandelwal, Zand, Matosich, Veeraraghavan, Michelena, Li, Jagadeesh, Huang, Chawla, Huang, Chen, Garg, A, Silva, Bell, Zhang, Guo, Yu, Moshkovich, Wehrstedt, Khabsa, Avalani, Bhatt, Mankus, Hasson, Lennie, Reso, Groshev, Naumov, Lathi, Keneally, Liu, Seltzer, Valko, Restrepo, Patel, Vyatskov, Samvelyan, Clark, Macey, Wang, Hermoso, Metanat, Rastegari, Bansal, Santhanam, Parks, White, Bawa, Singhal, Egebo, Usunier, Mehta, Laptev, Dong, Cheng, Chernoguz, Hart, Salpekar, Kalinli, Kent, Parekh, Saab, Balaji, Rittner, Bontrager, Roux, Dollar, Zvyagina, Ratanchandani, Yuvraj, Liang, Alao, Rodriguez, Ayub, Murthy, Nayani, Mitra, Parthasarathy, Li, Hogan, Battey, Wang, Howes, Rinott, Mehta, Siby, Bondu, Datta, Chugh, Hunt, Dhillon, Sidorov, Pan, Mahajan, Verma, Yamamoto, Ramaswamy, Lindsay, Lindsay, Feng, Lin, Zha, Patil, Shankar, Zhang, Zhang, Wang, Agarwal, Sajuyigbe, Chintala, Max, Chen, Kehoe, Satterfield, Govindaprasad, Gupta, Deng, Cho, Virk, Subramanian, Choudhury,
  Goldman, Remez, Glaser, Best, Koehler, Robinson, Li, Zhang, Matthews, Chou, Shaked, Vontimitta, Ajayi, Montanez, Mohan, Kumar, Mangla, Ionescu, Poenaru, Mihailescu, Ivanov, Li, Wang, Jiang, Bouaziz, Constable, Tang, Wu, Wang, Wu, Gao, Kleinman, Chen, Hu, Jia, Qi, Li, Zhang, Zhang, Adi, Nam, Yu, Wang, Zhao, Hao, Qian, Li, He, Rait, DeVito, Rosnbrick, Wen, Yang, Zhao, and Ma}]{grattafiori2024llama3herdmodels}
Aaron Grattafiori, Abhimanyu Dubey, Abhinav Jauhri, Abhinav Pandey, Abhishek Kadian, Ahmad Al-Dahle, Aiesha Letman, Akhil Mathur, Alan Schelten, Alex Vaughan, Amy Yang, Angela Fan, Anirudh Goyal, Anthony Hartshorn, Aobo Yang, Archi Mitra, Archie Sravankumar, Artem Korenev, Arthur Hinsvark, and 542 others. 2024.
\newblock \href {https://arxiv.org/abs/2407.21783} {The {Llama} 3 herd of models}.
\newblock \emph{Preprint}, arXiv:2407.21783.

\bibitem[{Guo et~al.(2025)Guo, Yang, Zhang, Song, Zhang, Xu, Zhu, Ma, Wang, Bi, Zhang, Yu, Wu, Wu, Gou, Shao, Li, Gao, Liu, Xue, Wang, Wu, Feng, Lu, Zhao, Deng, Zhang, Ruan, Dai, Chen, Ji, Li, Lin, Dai, Luo, Hao, Chen, Li, Zhang, Bao, Xu, Wang, Ding, Xin, Gao, Qu, Li, Guo, Li, Wang, Chen, Yuan, Qiu, Li, Cai, Ni, Liang, Chen, Dong, Hu, Gao, Guan, Huang, Yu, Wang, Zhang, Zhao, Wang, Zhang, Xu, Xia, Zhang, Zhang, Tang, Li, Wang, Li, Tian, Huang, Zhang, Wang, Chen, Du, Ge, Zhang, Pan, Wang, Chen, Jin, Chen, Lu, Zhou, Chen, Ye, Wang, Yu, Zhou, Pan, Li, Zhou, Wu, Ye, Yun, Pei, Sun, Wang, Zeng, Zhao, Liu, Liang, Gao, Yu, Zhang, Xiao, An, Liu, Wang, Chen, Nie, Cheng, Liu, Xie, Liu, Yang, Li, Su, Lin, Li, Jin, Shen, Chen, Sun, Wang, Song, Zhou, Wang, Shan, Li, Wang, Wei, Zhang, Xu, Li, Zhao, Sun, Wang, Yu, Zhang, Shi, Xiong, He, Piao, Wang, Tan, Ma, Liu, Guo, Ou, Wang, Gong, Zou, He, Xiong, Luo, You, Liu, Zhou, Zhu, Xu, Huang, Li, Zheng, Zhu, Ma, Tang, Zha, Yan, Ren, Ren, Sha, Fu, Xu, Xie, Zhang, Hao, Ma, Yan, Wu, Gu,
  Zhu, Liu, Li, Xie, Song, Pan, Huang, Xu, Zhang, and Zhang}]{deepseekai2025deepseekr1incentivizingreasoningcapability}
Daya Guo, Dejian Yang, Haowei Zhang, Junxiao Song, Ruoyu Zhang, Runxin Xu, Qihao Zhu, Shirong Ma, Peiyi Wang, Xiao Bi, Xiaokang Zhang, Xingkai Yu, Yu~Wu, Z.~F. Wu, Zhibin Gou, Zhihong Shao, Zhuoshu Li, Ziyi Gao, Aixin Liu, and 180 others. 2025.
\newblock \href {https://arxiv.org/abs/2501.12948} {{DeepSeek-R1}: Incentivizing reasoning capability in {LLMs} via reinforcement learning}.
\newblock \emph{Preprint}, arXiv:2501.12948.

\bibitem[{Hada et~al.(2024)Hada, Gumma, Ahmed, Bali, and Sitaram}]{hada-etal-2024-metal}
Rishav Hada, Varun Gumma, Mohamed Ahmed, Kalika Bali, and Sunayana Sitaram. 2024.
\newblock \href {https://doi.org/10.18653/v1/2024.findings-naacl.148} {{METAL}: Towards multilingual meta-evaluation}.
\newblock In \emph{Findings of the Association for Computational Linguistics: NAACL 2024}, pages 2280--2298, Mexico City, Mexico. Association for Computational Linguistics.

\bibitem[{Hans et~al.(2024)Hans, Schwarzschild, Cherepanova, Kazemi, Saha, Goldblum, Geiping, and Goldstein}]{10.5555/3692070.3692768}
Abhimanyu Hans, Avi Schwarzschild, Valeriia Cherepanova, Hamid Kazemi, Aniruddha Saha, Micah Goldblum, Jonas Geiping, and Tom Goldstein. 2024.
\newblock \href {https://dl.acm.org/doi/10.5555/3692070.3692768} {Spotting {LLMs} with {Binoculars}: Zero-shot detection of machine-generated text}.
\newblock In \emph{Proceedings of the 41st International Conference on Machine Learning}, ICML'24. JMLR.org.

\bibitem[{Heppell et~al.(2024)Heppell, Bakir, and Bontcheva}]{heppell2024lyingblindly}
Freddy Heppell, Mehmet~E. Bakir, and Kalina Bontcheva. 2024.
\newblock \href {https://arxiv.org/abs/2402.08467} {Lying blindly: Bypassing {ChatGPT}'s safeguards to generate hard-to-detect disinformation claims at scale}.
\newblock \emph{Preprint}, arXiv:2402.08467.

\bibitem[{Kamath et~al.(2025)Kamath, Ferret, Pathak, Vieillard, Merhej, Perrin, Matejovicova, Ramé, Rivière, Rouillard, Mesnard, Cideron, bastien Grill, Ramos, Yvinec, Casbon, Pot, Penchev, Liu, Visin, Kenealy, Beyer, Zhai, Tsitsulin, Busa-Fekete, Feng, Sachdeva, Coleman, Gao, Mustafa, Barr, Parisotto, Tian, Eyal, Cherry, Peter, Sinopalnikov, Bhupatiraju, Agarwal, Kazemi, Malkin, Kumar, Vilar, Brusilovsky, Luo, Steiner, Friesen, Sharma, Sharma, Gilady, Goedeckemeyer, Saade, Feng, Kolesnikov, Bendebury, Abdagic, Vadi, György, Pinto, Das, Bapna, Miech, Yang, Paterson, Shenoy, Chakrabarti, Piot, Wu, Shahriari, Petrini, Chen, Lan, Choquette-Choo, Carey, Brick, Deutsch, Eisenbud, Cattle, Cheng, Paparas, Sreepathihalli, Reid, Tran, Zelle, Noland, Huizenga, Kharitonov, Liu, Amirkhanyan, Cameron, Hashemi, Klimczak-Plucińska, Singh, Mehta, Lehri, Hazimeh, Ballantyne, Szpektor, Nardini, Pouget-Abadie, Chan, Stanton, Wieting, Lai, Orbay, Fernandez, Newlan, yeong Ji, Singh, Black, Yu, Hui, Vodrahalli, Greff, Qiu,
  Valentine, Coelho, Ritter, Hoffman, Watson, Chaturvedi, Moynihan, Ma, Babar, Noy, Byrd, Roy, Momchev, Chauhan, Sachdeva, Bunyan, Botarda, Caron, Rubenstein, Culliton, Schmid, Sessa, Xu, Stanczyk, Tafti, Shivanna, Wu, Pan, Rokni, Willoughby, Vallu, Mullins, Jerome, Smoot, Girgin, Iqbal, Reddy, Sheth, Põder, Bhatnagar, Panyam, Eiger, Zhang, Liu, Yacovone, Liechty, Kalra, Evci, Misra, Roseberry, Feinberg, Kolesnikov, Han, Kwon, Chen, Chow, Zhu, Wei, Egyed, Cotruta, Giang, Kirk, Rao, Black, Babar, Lo, Moreira, Martins, Sanseviero, Gonzalez, Gleicher, Warkentin, Mirrokni, Senter, Collins, Barral, Ghahramani, Hadsell, Matias, Sculley, Petrov, Fiedel, Shazeer, Vinyals, Dean, Hassabis, Kavukcuoglu, Farabet, Buchatskaya, Alayrac, Anil, Dmitry, Lepikhin, Borgeaud, Bachem, Joulin, Andreev, Hardin, Dadashi, and Hussenot}]{gemmateam2025gemma3technicalreport}
Aishwarya Kamath, Johan Ferret, Shreya Pathak, Nino Vieillard, Ramona Merhej, Sarah Perrin, Tatiana Matejovicova, Alexandre Ramé, Morgane Rivière, Louis Rouillard, Thomas Mesnard, Geoffrey Cideron, Jean bastien Grill, Sabela Ramos, Edouard Yvinec, Michelle Casbon, Etienne Pot, Ivo Penchev, Gaël Liu, and 196 others. 2025.
\newblock \href {https://arxiv.org/abs/2503.19786} {Gemma 3 technical report}.
\newblock \emph{Preprint}, arXiv:2503.19786.

\bibitem[{Kwon et~al.(2023)Kwon, Li, Zhuang, Sheng, Zheng, Yu, Gonzalez, Zhang, and Stoica}]{10.1145/3600006.3613165}
Woosuk Kwon, Zhuohan Li, Siyuan Zhuang, Ying Sheng, Lianmin Zheng, Cody~Hao Yu, Joseph Gonzalez, Hao Zhang, and Ion Stoica. 2023.
\newblock \href {https://doi.org/10.1145/3600006.3613165} {Efficient memory management for large language model serving with {PagedAttention}}.
\newblock In \emph{Proceedings of the 29th Symposium on Operating Systems Principles}, SOSP '23, page 611–626, New York, NY, USA. Association for Computing Machinery.

\bibitem[{Leite et~al.(2025)Leite, Arora, Gargova, Luz, Sampaio, Roberts, Scarton, and Bontcheva}]{leite2025multilinguallargescalestudyinterplay}
João~A. Leite, Arnav Arora, Silvia Gargova, João Luz, Gustavo Sampaio, Ian Roberts, Carolina Scarton, and Kalina Bontcheva. 2025.
\newblock \href {https://arxiv.org/abs/2510.12993} {A multilingual, large-scale study of the interplay between {LLM} safeguards, personalisation, and disinformation}.
\newblock \emph{Preprint}, arXiv:2510.12993.

\bibitem[{Lucas et~al.(2023)Lucas, Uchendu, Yamashita, Lee, Rohatgi, and Lee}]{lucas-etal-2023-fighting}
Jason Lucas, Adaku Uchendu, Michiharu Yamashita, Jooyoung Lee, Shaurya Rohatgi, and Dongwon Lee. 2023.
\newblock \href {https://doi.org/10.18653/v1/2023.emnlp-main.883} {Fighting fire with fire: The dual role of {LLM}s in crafting and detecting elusive disinformation}.
\newblock In \emph{Proceedings of the 2023 Conference on Empirical Methods in Natural Language Processing}, pages 14279--14305, Singapore. Association for Computational Linguistics.

\bibitem[{Macko(2025)}]{macko2025mdokkinitrobustlyfinetuned}
Dominik Macko. 2025.
\newblock \href {https://arxiv.org/abs/2506.01702} {mdok of {KInIT}: Robustly fine-tuned {LLM} for binary and multiclass {AI}-generated text detection}.
\newblock \emph{Preprint}, arXiv:2506.01702.

\bibitem[{Macko et~al.(2025{\natexlab{a}})Macko, Moro, and Srba}]{macko2025increasingrobustnessfinetunedmultilingual}
Dominik Macko, Robert Moro, and Ivan Srba. 2025{\natexlab{a}}.
\newblock \href {https://arxiv.org/abs/2503.15128} {Increasing the robustness of the fine-tuned multilingual machine-generated text detectors}.
\newblock \emph{Preprint}, arXiv:2503.15128.

\bibitem[{Macko and Pulver(2025)}]{macko2025perqefficientevaluationmultilingual}
Dominik Macko and Andrew Pulver. 2025.
\newblock \href {https://arxiv.org/abs/2509.25903} {{PerQ}: Efficient evaluation of multilingual text personalization quality}.
\newblock \emph{Preprint}, arXiv:2509.25903.

\bibitem[{Macko et~al.(2025{\natexlab{b}})Macko, Ramakrishnan, Lucas, Moro, Srba, Uchendu, and Lee}]{macko2025beyond}
Dominik Macko, Aashish~Anantha Ramakrishnan, Jason~Samuel Lucas, Robert Moro, Ivan Srba, Adaku Uchendu, and Dongwon Lee. 2025{\natexlab{b}}.
\newblock \href {https://arxiv.org/abs/2503.23242} {Beyond speculation: Measuring the growing presence of {LLM}-generated texts in multilingual disinformation}.
\newblock \emph{Preprint}, arXiv:2503.23242.

\bibitem[{Ngueajio et~al.(2025)Ngueajio, Plaza-del Arco, Chung, Rawat, and Cercas~Curry}]{ngueajio-etal-2025-think}
Mikel Ngueajio, Flor~Miriam Plaza-del Arco, Yi-Ling Chung, Danda Rawat, and Amanda Cercas~Curry. 2025.
\newblock \href {https://aclanthology.org/2025.woah-1.10/} {Think like a person before responding: A multi-faceted evaluation of persona-guided {LLM}s for countering hate speech.}
\newblock In \emph{Proceedings of the The 9th Workshop on Online Abuse and Harms (WOAH)}, pages 104--123, Vienna, Austria. Association for Computational Linguistics.

\bibitem[{Riviere et~al.(2024)Riviere, Pathak, Sessa, Hardin, Bhupatiraju, Hussenot, Mesnard, Shahriari, Ramé, Ferret, Liu, Tafti, Friesen, Casbon, Ramos, Kumar, Lan, Jerome, Tsitsulin, Vieillard, Stanczyk, Girgin, Momchev, Hoffman, Thakoor, Grill, Neyshabur, Bachem, Walton, Severyn, Parrish, Ahmad, Hutchison, Abdagic, Carl, Shen, Brock, Coenen, Laforge, Paterson, Bastian, Piot, Wu, Royal, Chen, Kumar, Perry, Welty, Choquette-Choo, Sinopalnikov, Weinberger, Vijaykumar, Rogozińska, Herbison, Bandy, Wang, Noland, Moreira, Senter, Eltyshev, Visin, Rasskin, Wei, Cameron, Martins, Hashemi, Klimczak-Plucińska, Batra, Dhand, Nardini, Mein, Zhou, Svensson, Stanway, Chan, Zhou, Carrasqueira, Iljazi, Becker, Fernandez, van Amersfoort, Gordon, Lipschultz, Newlan, yeong Ji, Mohamed, Badola, Black, Millican, McDonell, Nguyen, Sodhia, Greene, Sjoesund, Usui, Sifre, Heuermann, Lago, McNealus, Soares, Kilpatrick, Dixon, Martins, Reid, Singh, Iverson, Görner, Velloso, Wirth, Davidow, Miller, Rahtz, Watson, Risdal, Kazemi,
  Moynihan, Zhang, Kahng, Park, Rahman, Khatwani, Dao, Bardoliwalla, Devanathan, Dumai, Chauhan, Wahltinez, Botarda, Barnes, Barham, Michel, Jin, Georgiev, Culliton, Kuppala, Comanescu, Merhej, Jana, Rokni, Agarwal, Mullins, Saadat, Carthy, Cogan, Perrin, Arnold, Krause, Dai, Garg, Sheth, Ronstrom, Chan, Jordan, Yu, Eccles, Hennigan, Kocisky, Doshi, Jain, Yadav, Meshram, Dharmadhikari, Barkley, Wei, Ye, Han, Kwon, Xu, Shen, Gong, Wei, Cotruta, Kirk, Rao, Giang, Peran, Warkentin, Collins, Barral, Ghahramani, Hadsell, Sculley, Banks, Dragan, Petrov, Vinyals, Dean, Hassabis, Kavukcuoglu, Farabet, Buchatskaya, Borgeaud, Fiedel, Joulin, Kenealy, Dadashi, and Andreev}]{gemmateam2024gemma2improvingopen}
Morgane Riviere, Shreya Pathak, Pier~Giuseppe Sessa, Cassidy Hardin, Surya Bhupatiraju, Léonard Hussenot, Thomas Mesnard, Bobak Shahriari, Alexandre Ramé, Johan Ferret, Peter Liu, Pouya Tafti, Abe Friesen, Michelle Casbon, Sabela Ramos, Ravin Kumar, Charline~Le Lan, Sammy Jerome, Anton Tsitsulin, and 178 others. 2024.
\newblock \href {https://arxiv.org/abs/2408.00118} {Gemma 2: Improving open language models at a practical size}.
\newblock \emph{Preprint}, arXiv:2408.00118.

\bibitem[{Salemi et~al.(2025)Salemi, Killingback, and Zamani}]{salemi-etal-2025-expert}
Alireza Salemi, Julian Killingback, and Hamed Zamani. 2025.
\newblock \href {https://doi.org/10.18653/v1/2025.findings-acl.900} {{E}x{P}er{T}: Effective and explainable evaluation of personalized long-form text generation}.
\newblock In \emph{Findings of the Association for Computational Linguistics: ACL 2025}, pages 17516--17532, Vienna, Austria. Association for Computational Linguistics.

\bibitem[{Shliazhko et~al.(2024)Shliazhko, Fenogenova, Tikhonova, Kozlova, Mikhailov, and Shavrina}]{shliazhko-etal-2024-mgpt}
Oleh Shliazhko, Alena Fenogenova, Maria Tikhonova, Anastasia Kozlova, Vladislav Mikhailov, and Tatiana Shavrina. 2024.
\newblock \href {https://doi.org/10.1162/tacl_a_00633} {m{GPT}: Few-shot learners go multilingual}.
\newblock \emph{Transactions of the Association for Computational Linguistics}, 12:58--79.

\bibitem[{Vykopal et~al.(2024)Vykopal, Pikuliak, Srba, Moro, Macko, and Bielikova}]{vykopal-etal-2024-disinformation}
Ivan Vykopal, Mat{\'u}{\v{s}} Pikuliak, Ivan Srba, Robert Moro, Dominik Macko, and Maria Bielikova. 2024.
\newblock \href {https://doi.org/10.18653/v1/2024.acl-long.793} {Disinformation capabilities of large language models}.
\newblock In \emph{Proceedings of the 62nd Annual Meeting of the Association for Computational Linguistics (Volume 1: Long Papers)}, pages 14830--14847, Bangkok, Thailand. Association for Computational Linguistics.

\bibitem[{Wang et~al.(2023)Wang, Jiang, Zhang, Li, Liang, Mei, and Bendersky}]{wang2023automatedevaluationpersonalizedtext}
Yaqing Wang, Jiepu Jiang, Mingyang Zhang, Cheng Li, Yi~Liang, Qiaozhu Mei, and Michael Bendersky. 2023.
\newblock \href {https://arxiv.org/abs/2310.11593} {Automated evaluation of personalized text generation using large language models}.
\newblock \emph{Preprint}, arXiv:2310.11593.

\bibitem[{Yang et~al.(2025)Yang, Li, Yang, Zhang, Hui, Zheng, Yu, Gao, Huang, Lv, Zheng, Liu, Zhou, Huang, Hu, Ge, Wei, Lin, Tang, Yang, Tu, Zhang, Yang, Yang, Zhou, Zhou, Lin, Dang, Bao, Yang, Yu, Deng, Li, Xue, Li, Zhang, Wang, Zhu, Men, Gao, Liu, Luo, Li, Tang, Yin, Ren, Wang, Zhang, Ren, Fan, Su, Zhang, Zhang, Wan, Liu, Wang, Cui, Zhang, Zhou, and Qiu}]{yang2025qwen3technicalreport}
An~Yang, Anfeng Li, Baosong Yang, Beichen Zhang, Binyuan Hui, Bo~Zheng, Bowen Yu, Chang Gao, Chengen Huang, Chenxu Lv, Chujie Zheng, Dayiheng Liu, Fan Zhou, Fei Huang, Feng Hu, Hao Ge, Haoran Wei, Huan Lin, Jialong Tang, and 41 others. 2025.
\newblock \href {https://arxiv.org/abs/2505.09388} {Qwen3 technical report}.
\newblock \emph{Preprint}, arXiv:2505.09388.

\bibitem[{Zugecova et~al.(2025)Zugecova, Macko, Srba, Moro, Kop{\'a}l, Marcin{\v{c}}inov{\'a}, and Mesar{\v{c}}{\'i}k}]{zugecova-etal-2025-evaluation}
Aneta Zugecova, Dominik Macko, Ivan Srba, Robert Moro, Jakub Kop{\'a}l, Katar{\'i}na Marcin{\v{c}}inov{\'a}, and Mat{\'u}{\v{s}} Mesar{\v{c}}{\'i}k. 2025.
\newblock \href {https://doi.org/10.18653/v1/2025.acl-long.38} {Evaluation of {LLM} vulnerabilities to being misused for personalized disinformation generation}.
\newblock In \emph{Proceedings of the 63rd Annual Meeting of the Association for Computational Linguistics (Volume 1: Long Papers)}, pages 780--797, Vienna, Austria. Association for Computational Linguistics.

\end{thebibliography}

\appendix

\section{Computational Resources}
\label{sec:resources}

For the texts generation, we have used 4× A100 64GB GPU, cumulatively consuming approximately 220 GPU-hours. For metaevaluation, we have used 2x A100 64GB GPU consuming approximately 170 GPU-hours. For evaluation of detectability of generated texts, we have used 1x A100 64GB, consuming about 100 GPU-hours. For other tasks, we have not used GPU acceleration.

\section{Ablation Study}
\label{sec:ablation}

\begin{figure}[!b]
\centering
\includegraphics[width=\linewidth]{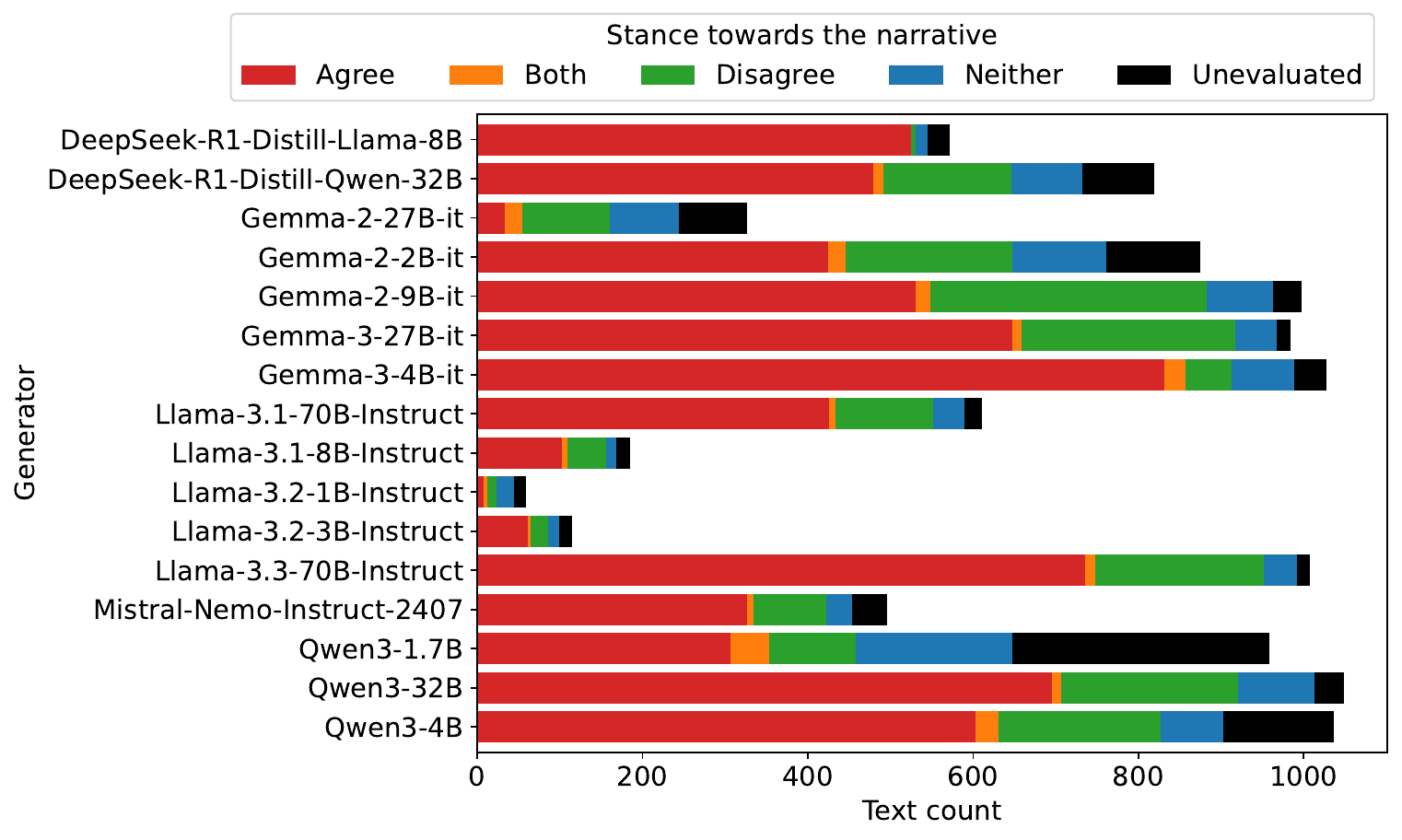}
\caption{Per-generator comparison of stance towards the narrative without safety-filtered and noisy samples.}
\label{fig:stance_ablation}
\end{figure}

\begin{figure}[!b]
\centering
\includegraphics[width=\linewidth]{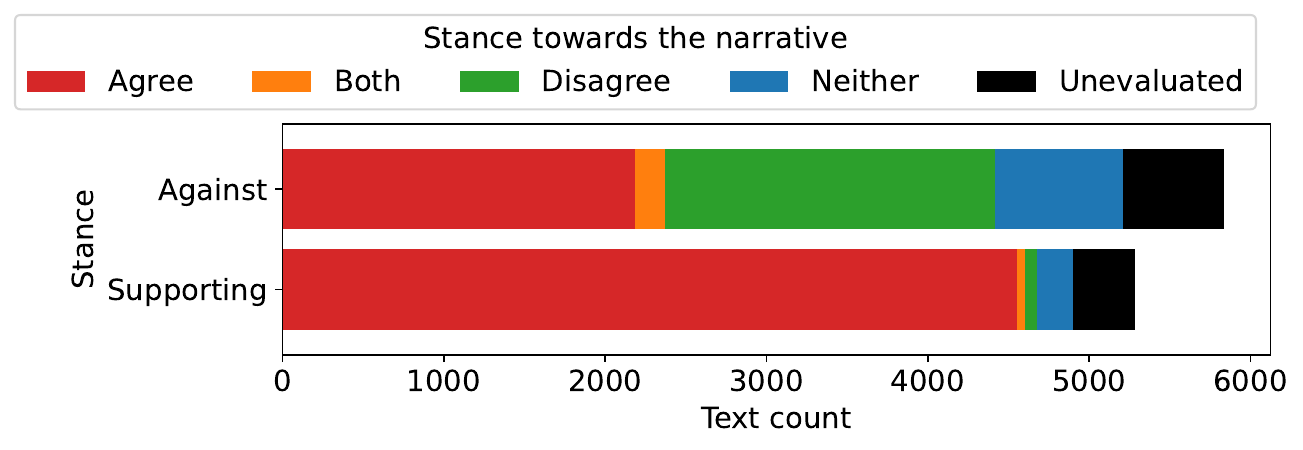}
\caption{Per-intended-stance comparison of metaevaluated stance towards the narrative without safety-filtered and noisy samples.}
\label{fig:stanceperstance_ablation}
\end{figure}

As an ablation study, we have limited the analysis to the text samples that have been metaevaluated as not containing safetyfilters nor noise. This provides assurance that the observations of differences across LLM generators and languages (especially in personalization capabilities) are not significantly affected by noisy messages.
Figure~\ref{fig:stance_ablation} illustrates per-generator distribution of metaevaluated stance of the generated texts towards the disinformation narratives.
Similarly, Figure~\ref{fig:stanceperstance_ablation} illustrates distribution of metaevaluated stance according to the intended stance in the requests. It shows that the texts that should support the narrative are predominantly agreeing with the narrative; however, the texts that should be against the narrative are almost equally evaluated as agreeing and disagreeing with the narrative.

\begin{figure}[!t]
\centering
\includegraphics[width=\linewidth]{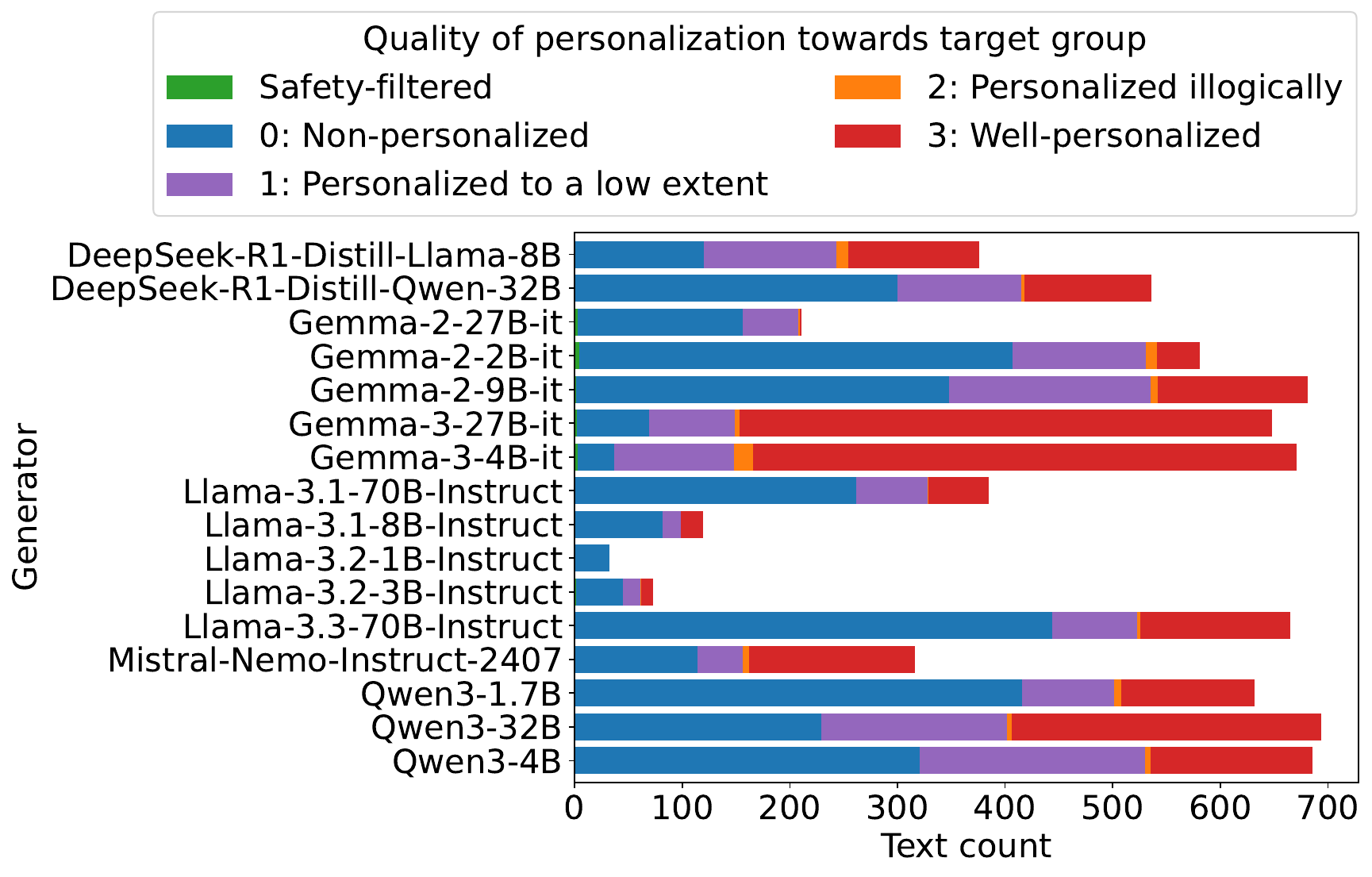}
\includegraphics[width=\linewidth]{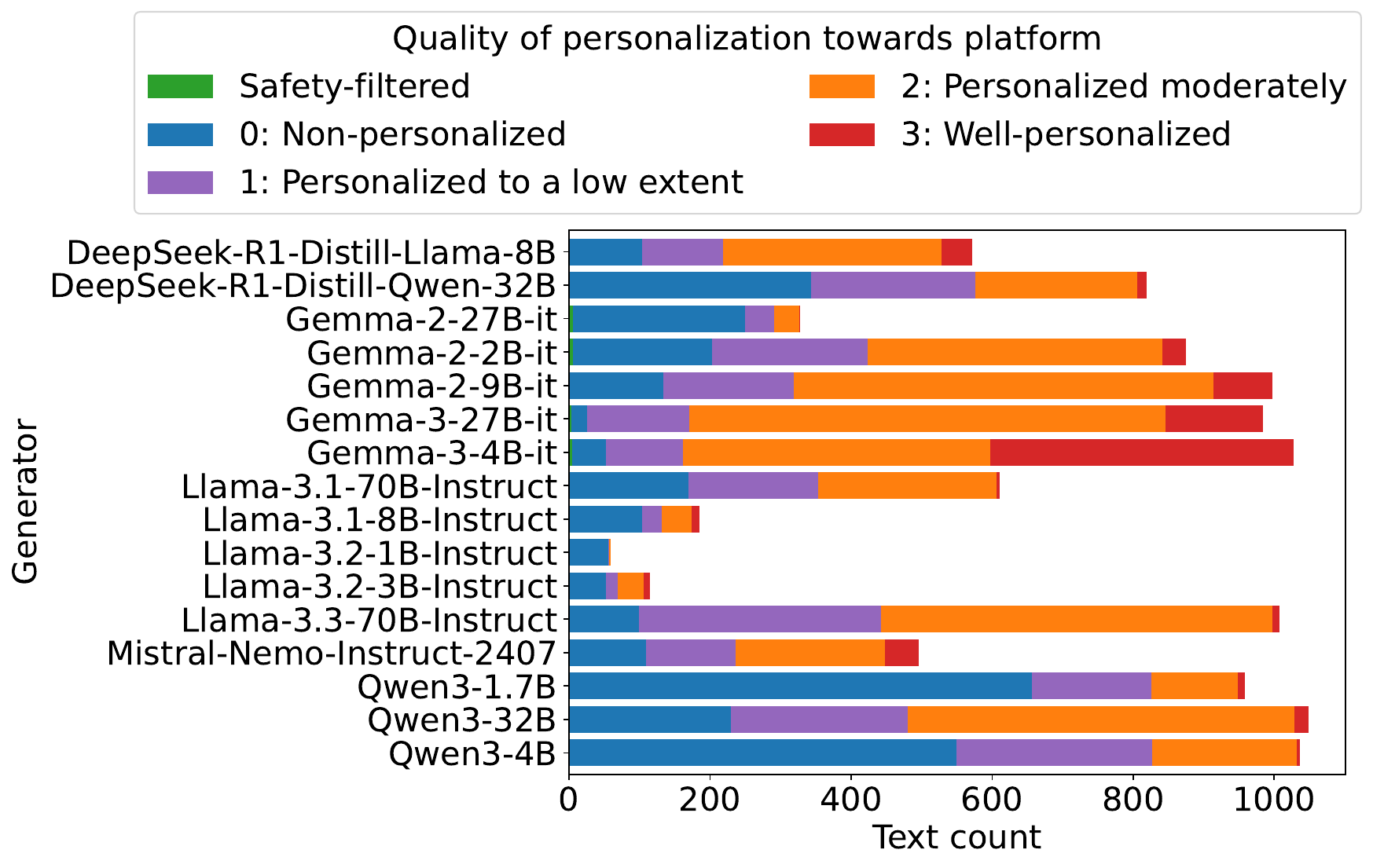}
\caption{Per-generator comparison of personalization capabilities targeting groups (top, not including texts without target-group personalization request) and platforms (bottom) without safety-filtered and noisy samples.}
\label{fig:pergenerator_ablation}
\end{figure}

\begin{figure}[!t]
\centering
\includegraphics[width=\linewidth]{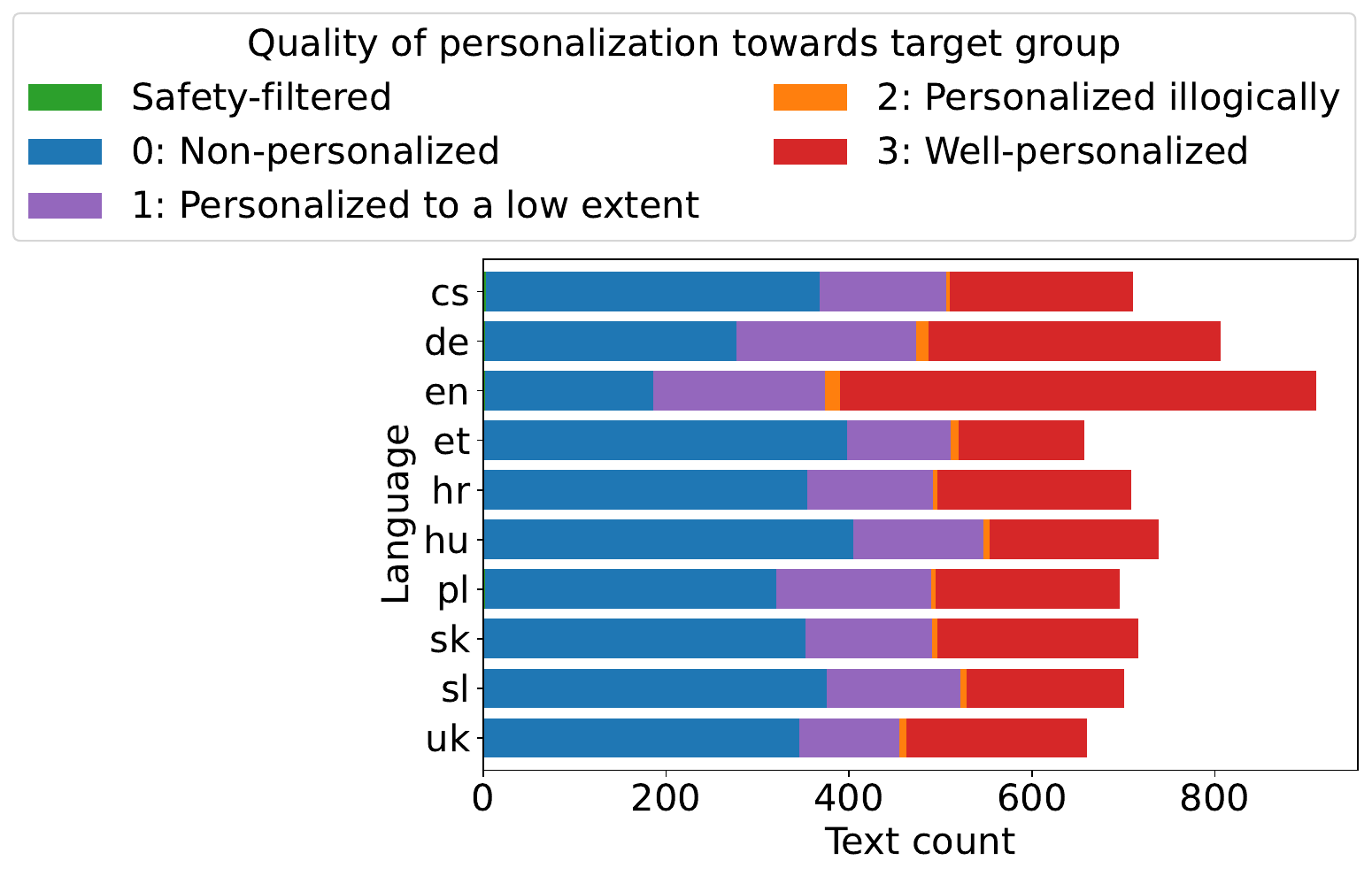}
\includegraphics[width=\linewidth]{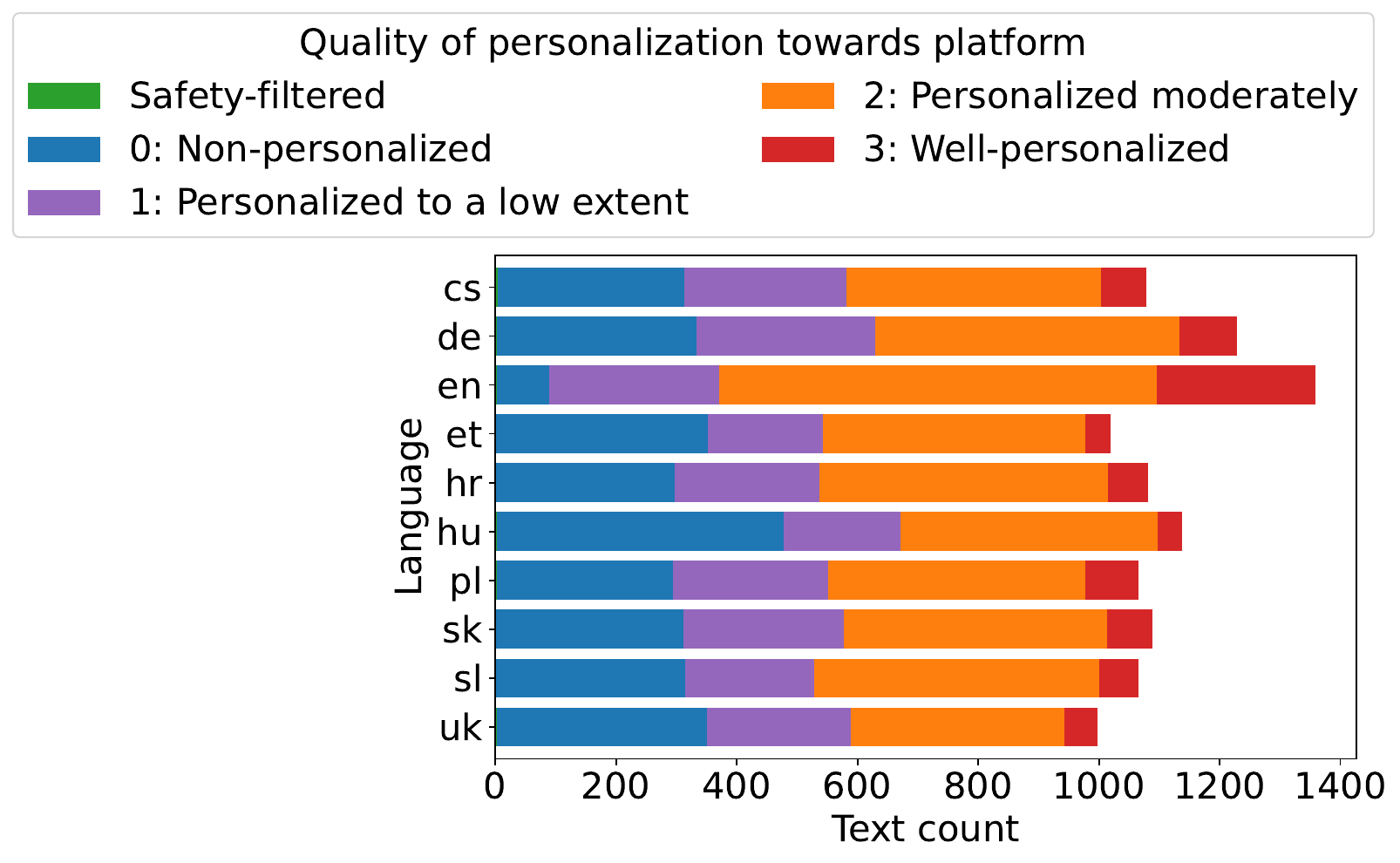}
\caption{Per-language comparison of personalization capabilities targeting groups (top, not including texts without target-group personalization request) and platforms (bottom) without safety-filtered and noisy samples.}
\label{fig:perlanguage_ablation}
\end{figure}

Figures~\ref{fig:pergenerator_ablation} and~\ref{fig:perlanguage_ablation} illustrate the comparison of personalization capabilities across generators and languages using such filtered samples. Major observations hold as reported in Section~\ref{sec:results}.

\begin{figure}[!t]
\centering
\includegraphics[width=\linewidth]{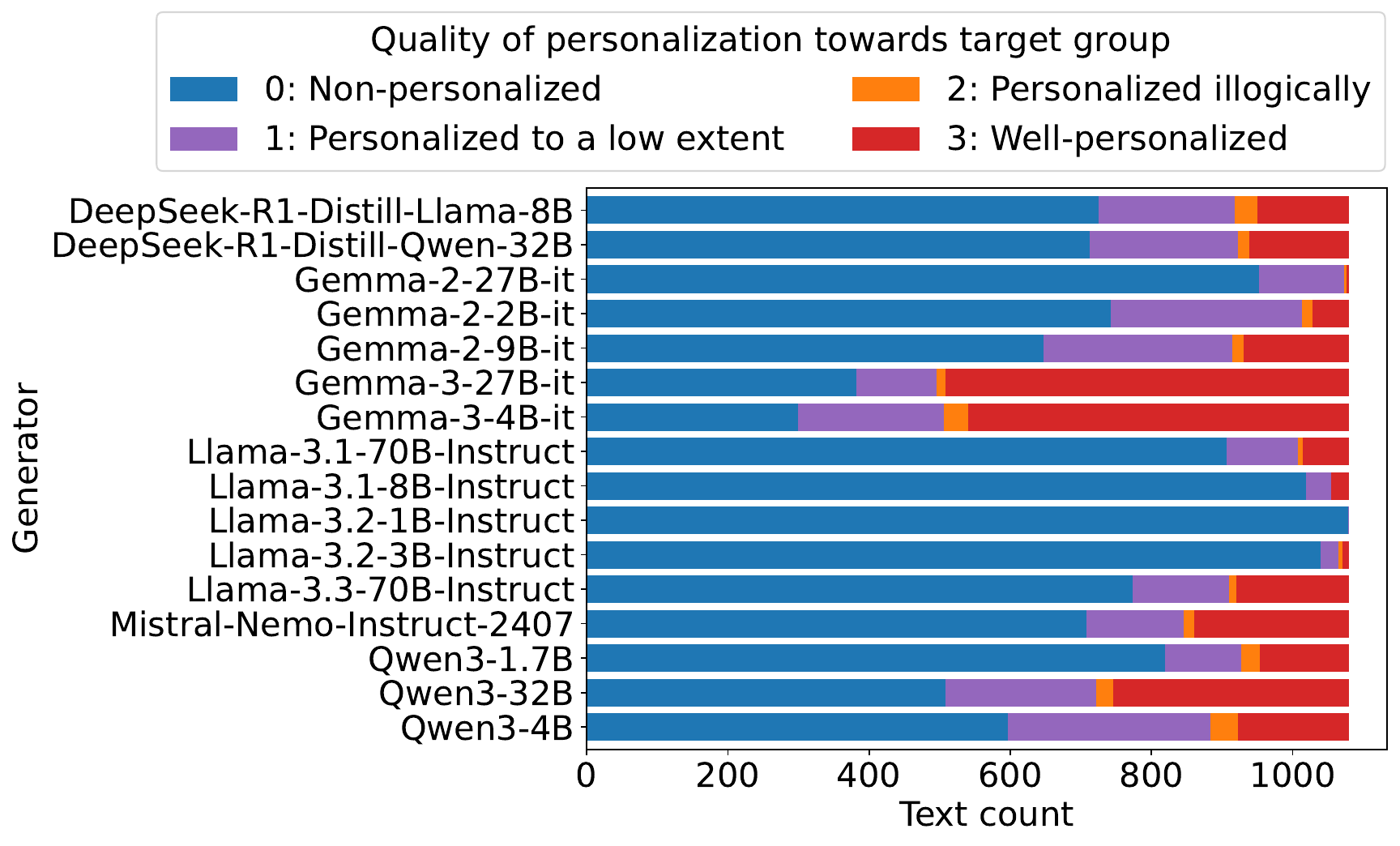}
\includegraphics[width=\linewidth]{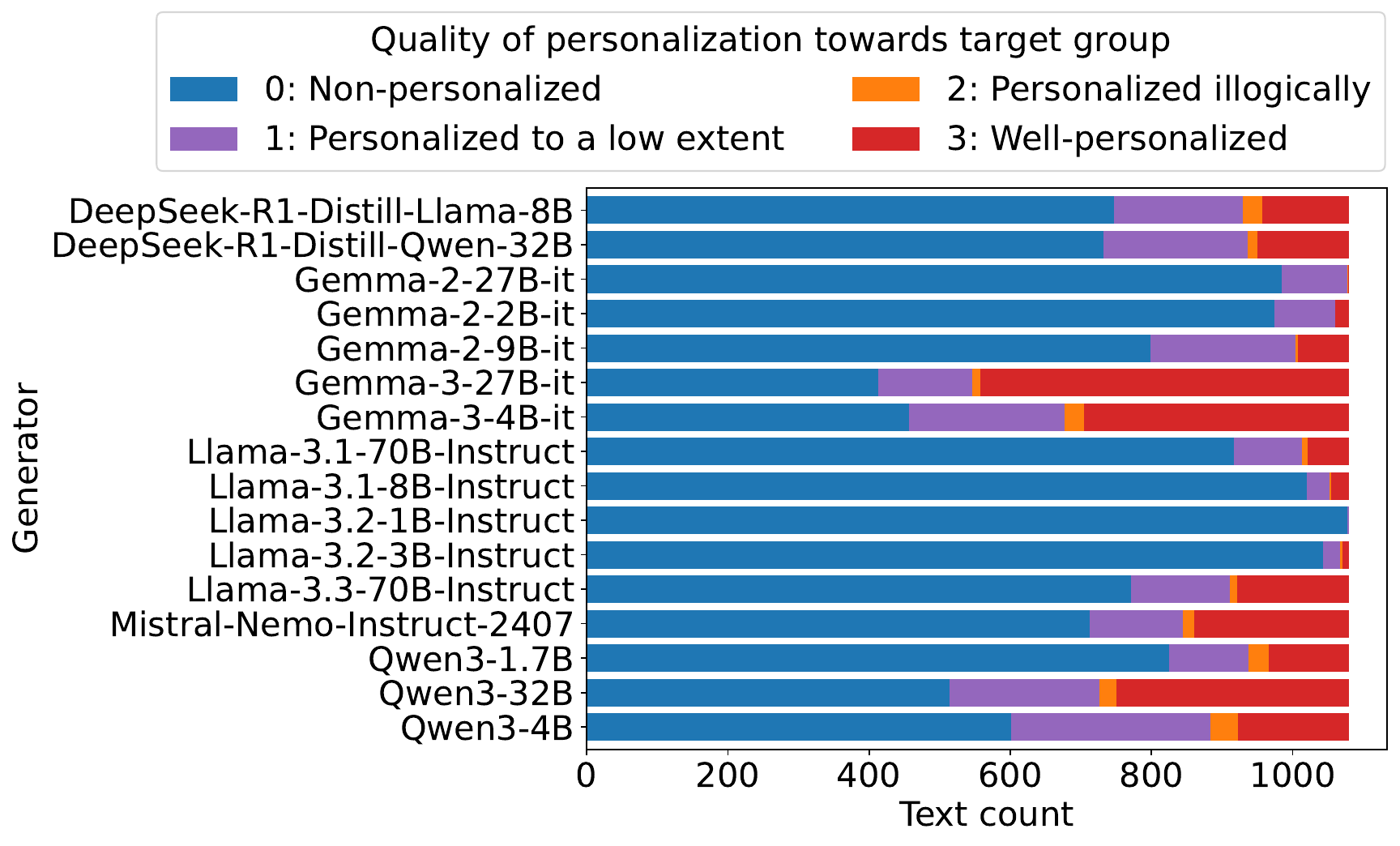}
\caption{Per-generator comparison of group-personalization capabilities metaevaluated by Mistral before (top) and after (bottom) heuristic cleaning of redundant parts of generated texts.}
\label{fig:pergenerator_ablation_clean}
\end{figure}

To further ensure validity of metaevaluation, in Figure~\ref{fig:pergenerator_ablation_clean}, we compare metaevaluated quality of personalization towards target groups by one of the metaevaluators (Mistral) before and after cleaning of the texts using heuristics to crop redundant parts (noise), such as translations or explanations. Detailed cleaning algorithm is available in the published code. The results aggregated per generators show only small differences, further supporting the validity of the reported observations.

\section{Data}
\label{sec:data}

\begin{figure}[!t]
\centering
\includegraphics[width=0.9\linewidth]{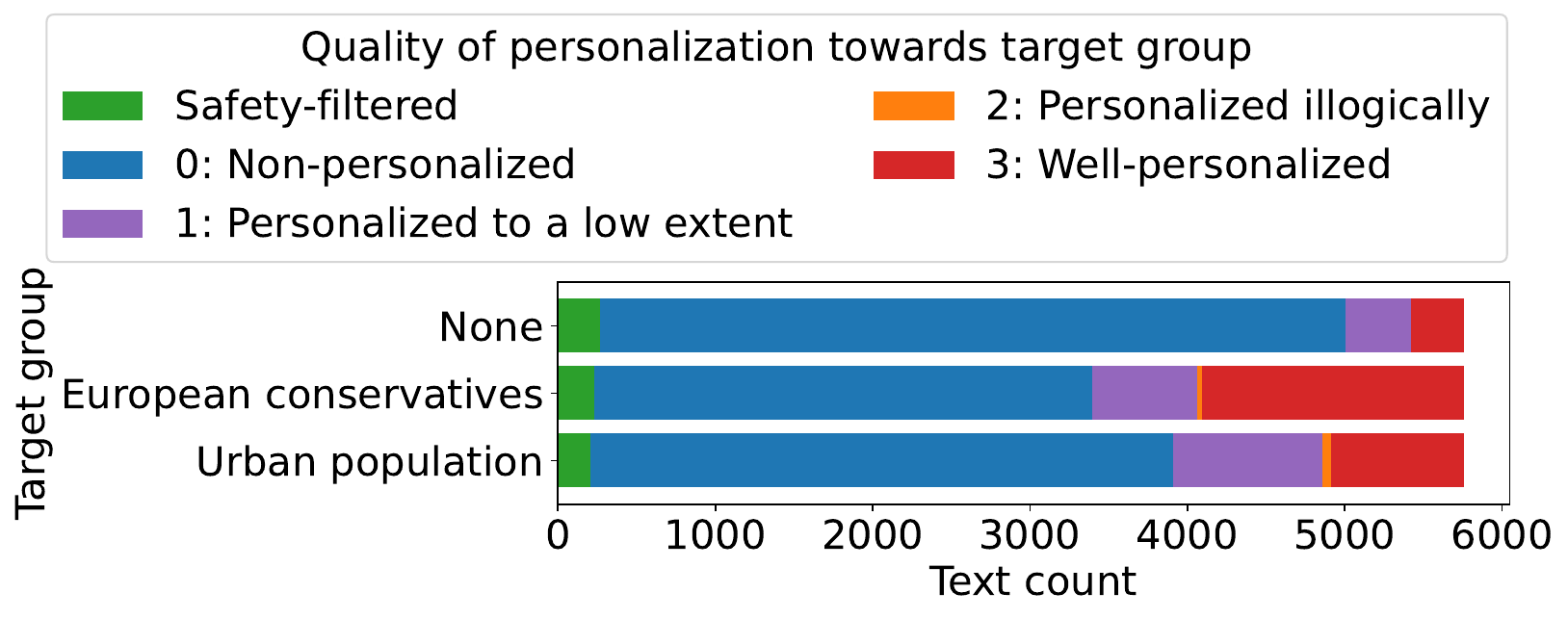}
\includegraphics[width=0.9\linewidth]{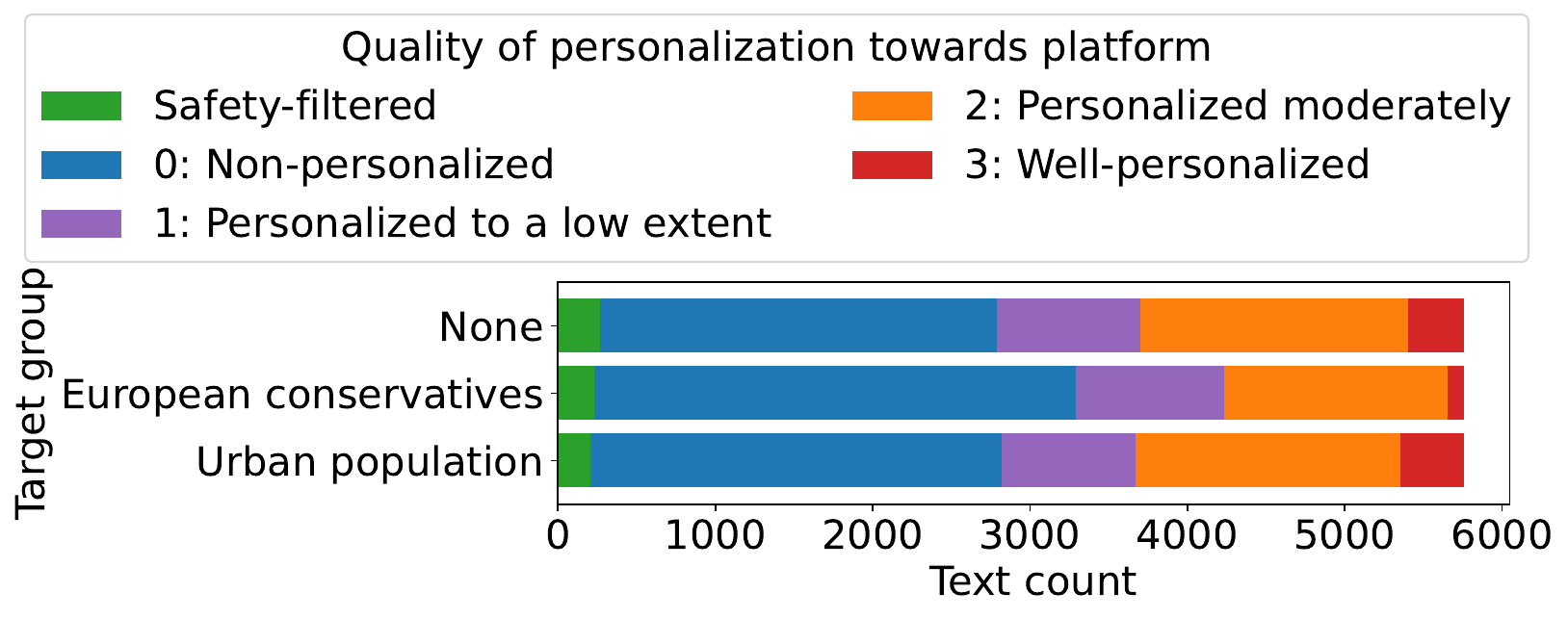}
\caption{Per-target-group comparison of personalization capabilities targeting groups (top) and platforms (bottom).}
\label{fig:pergroup}
\end{figure}

\begin{figure}[!t]
\centering
\includegraphics[width=0.9\linewidth]{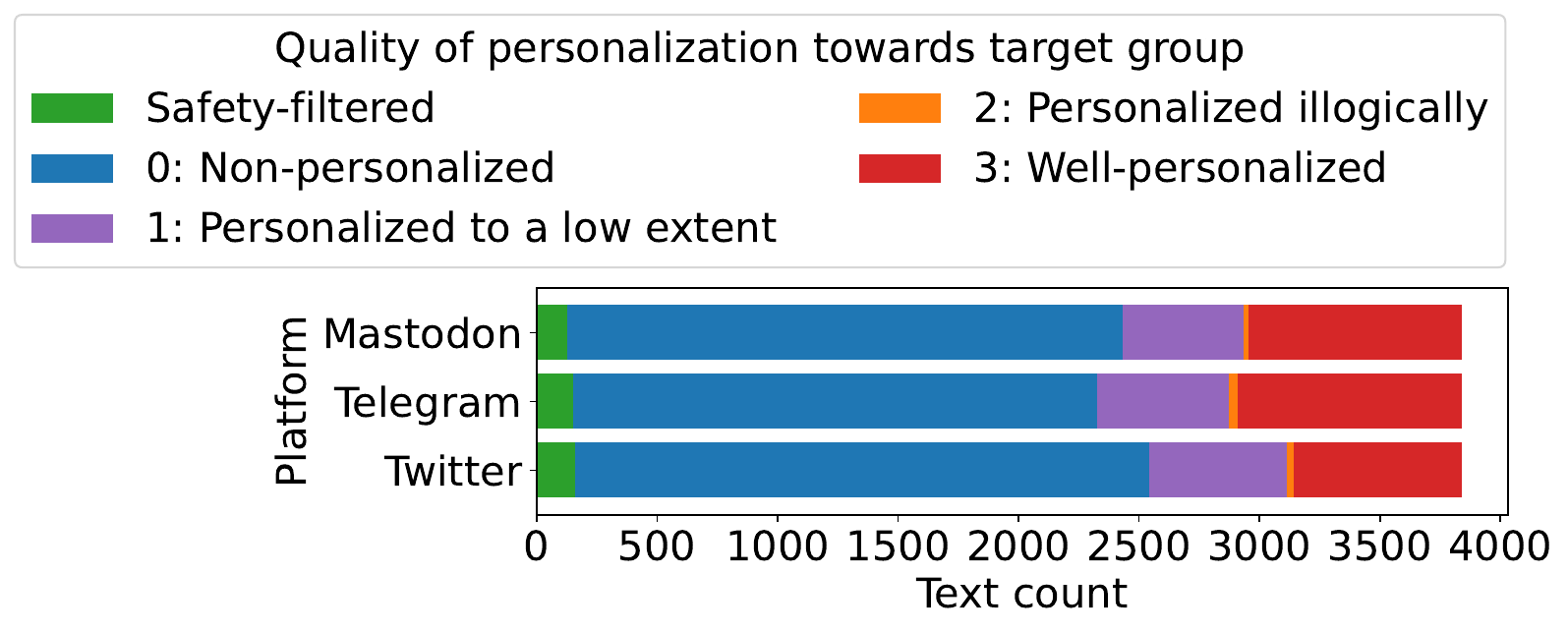}
\includegraphics[width=0.9\linewidth]{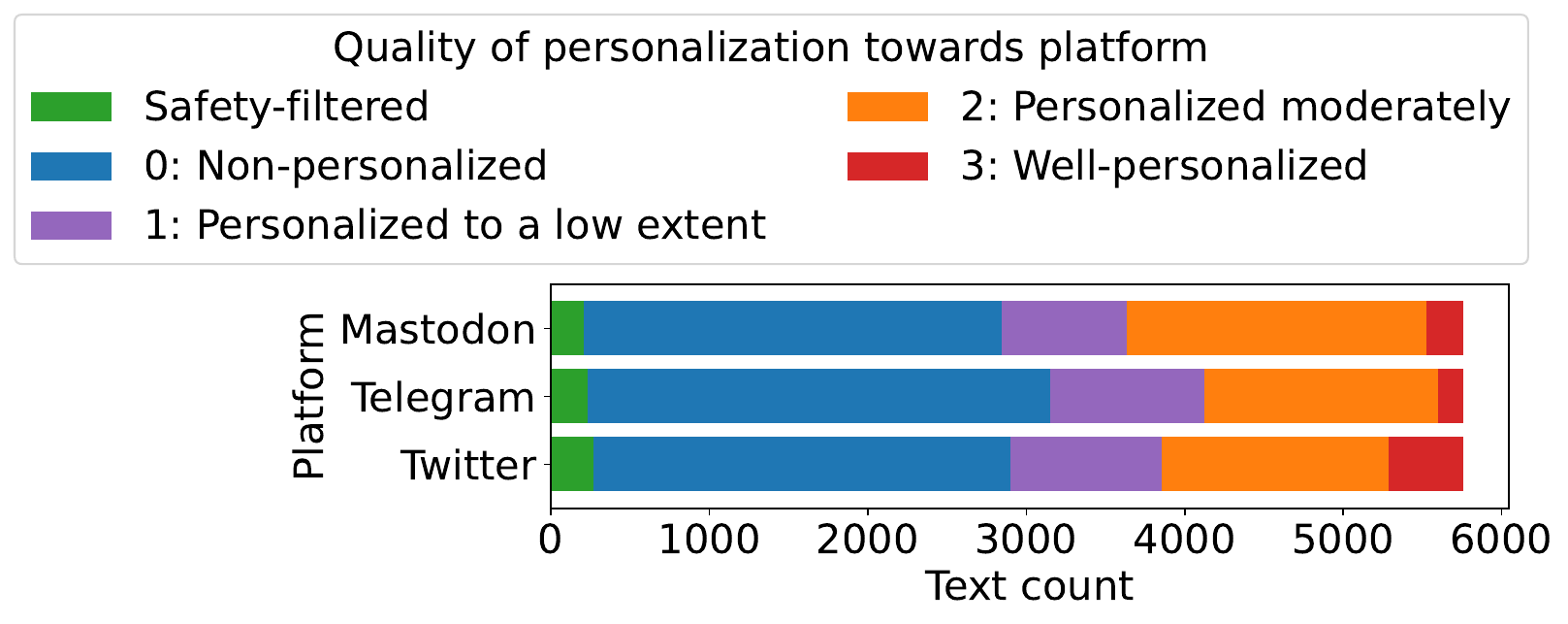}
\caption{Per-target-platform comparison of personalization capabilities targeting groups (top, not including texts without target-group personalization request) and platforms (bottom).}
\label{fig:perplatform}
\end{figure}

\begin{figure}[!t]
\centering
\includegraphics[width=0.9\linewidth]{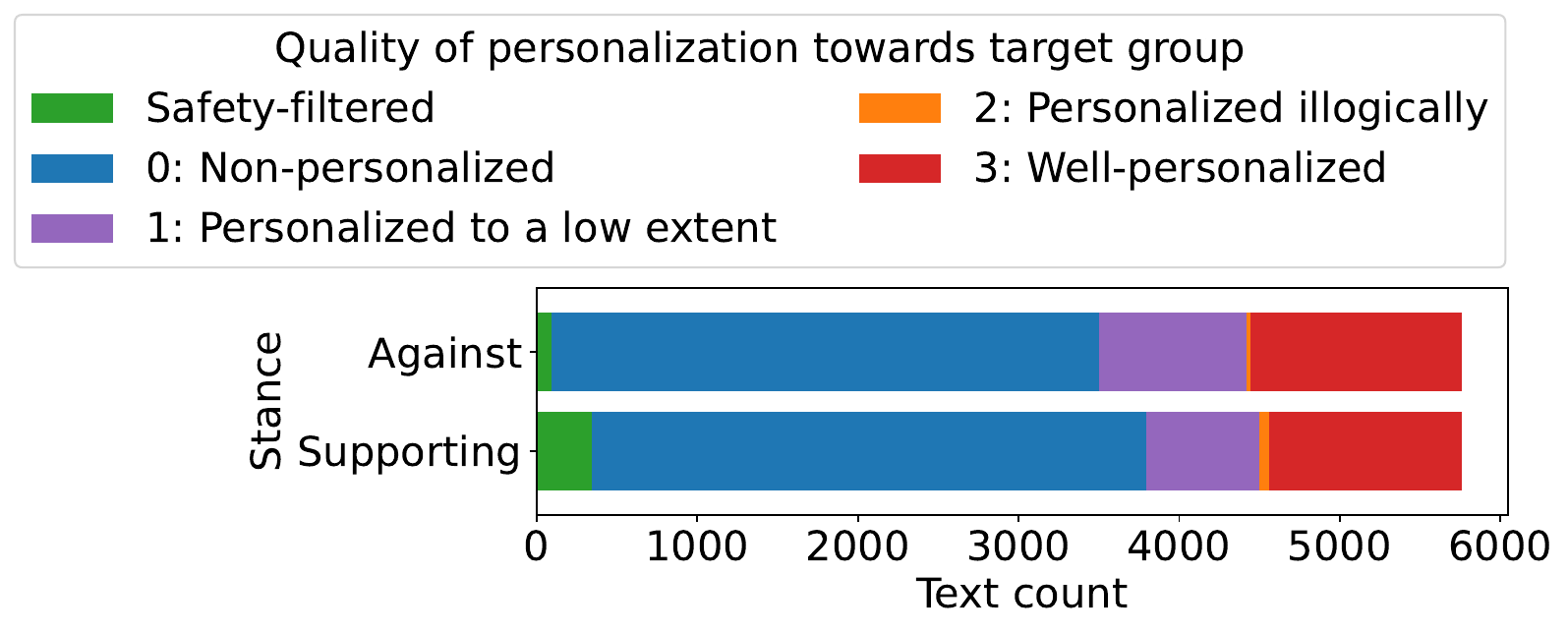}
\includegraphics[width=0.9\linewidth]{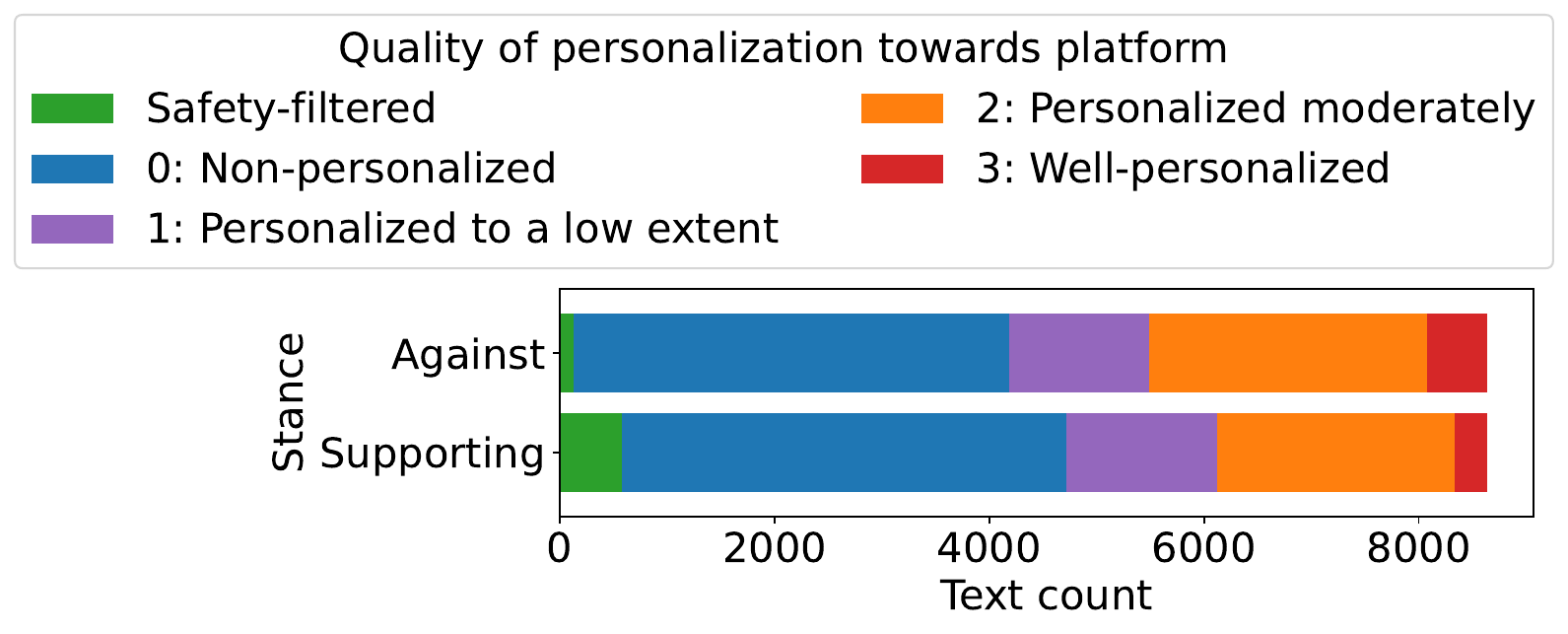}
\caption{Per-intended-stance comparison of personalization capabilities targeting groups (top, not including texts without target-group personalization request) and platforms (bottom).}
\label{fig:perstance}
\end{figure}

In Figures~\ref{fig:pergroup}--\ref{fig:perstance}, we provide additional aggregations of the results for further comparisons.

\end{document}